\documentclass{article} 
\usepackage[]{iclr2026_conference,times}

\usepackage{amsmath,amsfonts,bm}

\def\eqref#1{equation~\ref{#1}}

\def\1{\bm{1}}

\DeclareMathAlphabet{\mathsfit}{\encodingdefault}{\sfdefault}{m}{sl}
\SetMathAlphabet{\mathsfit}{bold}{\encodingdefault}{\sfdefault}{bx}{n}

\usepackage{hyperref}
\usepackage{url}
\usepackage{amsmath}
\usepackage{amssymb}
\usepackage{amsthm}
\usepackage{xcolor}
\usepackage{amsfonts}
\usepackage{mathtools}
\usepackage{makecell} 
\usepackage{multirow}
\usepackage{graphicx}
\theoremstyle{plain}
\newtheorem{theorem}{Theorem}[section]

\theoremstyle{definition}

\theoremstyle{remark}
\newtheorem{remark}[theorem]{Remark}

\usepackage{algorithm}
\usepackage{algorithmic}
\usepackage[algo2e]{algorithm2e}
\usepackage[utf8]{inputenc}
\usepackage[cjk]{kotex}
\usepackage{wrapfig}
\usepackage{subfigure}
\usepackage{multirow}
\usepackage{pbox}
\usepackage{graphicx}
\usepackage{bm}
\usepackage{siunitx}
\usepackage{booktabs}
\usepackage{lipsum}
\usepackage{caption}
\usepackage{fixltx2e}
\usepackage{cancel}
\usepackage{colortbl}
\usepackage{enumitem}
\usepackage{xfrac}
\usepackage{bbding}
\usepackage{nicefrac}
\usepackage{microtype}

\usepackage[noabbrev,capitalize,nameinlink]{cleveref}
\Crefname{figure}{Fig.}{Figs.}
\Crefname{equation}{Eq.}{Eqs.}
\newcommand{\std}[1]{\tiny{\ensuremath{\pm}}\text{\tiny #1}}

\title{Learning Posterior Predictive Distributions for Node Classification from Synthetic Graph Priors}

\author{Jeongwhan Choi\thanks{Equal contribution.}\;\;\textsuperscript{\href{mailto:jeongwhan.choi@kaist.ac.kr}{\Envelope}}, Jongwoo Kim\footnotemark[1],\;\; Woosung Kang,\;\; Noseong Park\thanks{Corresponding Author.} \\
KAIST\\
Daejeon, Republic of Korea
}

\iclrfinalcopy 
\begin{document}

\maketitle

\begin{abstract}
One of the most challenging problems in graph machine learning is generalizing across graphs with diverse properties. 
Graph neural networks (GNNs) face a fundamental limitation: they require separate training for each new graph, preventing universal generalization across diverse graph datasets.
A critical challenge facing GNNs lies in their reliance on labeled training data for each individual graph, a requirement that hinders the capacity for universal node classification due to the heterogeneity inherent in graphs --- differences in homophily levels, community structures, and feature distributions across datasets. Inspired by the success of large language models (LLMs) that achieve in-context learning through massive-scale pre-training on diverse datasets, we introduce NodePFN. This universal node classification method generalizes to arbitrary graphs without graph-specific training. NodePFN learns posterior predictive distributions (PPDs) by training only on thousands of synthetic graphs generated from carefully designed priors. Our synthetic graph generation covers real-world graphs through the use of random networks with controllable homophily levels and structural causal models for complex feature-label relationships. We develop a dual-branch architecture combining context-query attention mechanisms with local message passing to enable graph-aware in-context learning. Extensive evaluation on 23 benchmarks demonstrates that a single pre-trained NodePFN achieves 71.27\% average accuracy. These results validate that universal graph learning patterns can be effectively learned from synthetic priors, establishing a new paradigm for generalization in node classification.
\end{abstract}

\section{Introduction} 
Graph neural networks (GNNs) have achieved success in tasks on graph-structured data prevalent in chemistry~\citep{Gilmer2017chemi,hamilton2020graph}, recommender systems~\citep{Rex2018pinsage,He20LightGCN}, biology~\citep{bongini2022biognn}, social sciences~\citep{kipf2017GCN,qiu2018deepinf}, etc, by learning to aggregate neighborhood information through message passing. However, GNNs still have the limitation that, for node classification~\citep{kipf2017GCN,bresson2017GatedGCN,hamilton2017graphSAGE,Klicpera2019APPNP,zhou2020graph,choi2023gread,luan2023graph,kim2025leveraging,choi2025fn}, separate GNN models must be trained for the labeled nodes of each new graph.
This dependence on graph-specific training makes generalization across graphs with different properties challenging.
The core issue is that real-world graphs exhibit vastly different structural properties --- varying homophily levels, community structures and features, and degree distributions among datasets.
GNNs struggle to handle this diversity without dataset-specific training.

The success of foundation models, particularly large language models (LLMs)~\citep{brown2020language,touvron2023llama,achiam2023gpt} comes from their training paradigm of learning generalizable patterns from massive and diverse datasets. This enables these models to perform in-context learning, adapting to new datasets without parameter updates by learned patterns during pre-training.
In a manner analogous to the capacity of LLMs to adapt to new samples with only context examples, we propose a graph model that performs node classification on arbitrary graph datasets. This implies that \emph{a single pre-trained model} could perform node classification on arbitrary graph datasets without needing to be trained specifically for that dataset.

Recent studies have explored applying LLMs to graphs~\citep{li2024graph,chen2024text,chen2024llaga,li2024glbench,tang2024graphgpt,liu2024ofa}. 
However, LLMs, primarily trained on textual data, are better suited for capturing semantic content rather than learning the structural patterns that govern node classification on diverse graph topologies.
While \citet{zhao2025graphany} introduce a fully inductive framework, it still requires training on specific source datasets, with performance varying significantly based on the training dataset choice.

\begin{figure}[t]
    \centering
    \resizebox{0.95\linewidth}{!}{%
        \begin{tabular}{cc}
            \subfigure[Standard GNNs]{\includegraphics[width=0.35\textwidth]{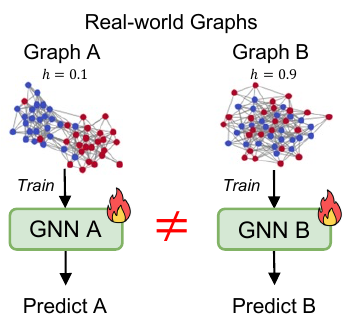}\label{fig:fn-a}} 
            \hfill
            \subfigure[Our proposed NodePFN]{\includegraphics[width=0.62\textwidth]{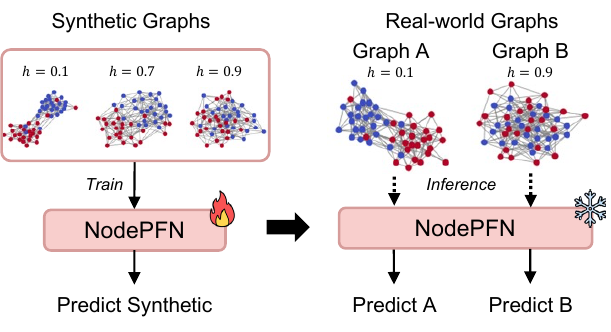}\label{fig:fn-b}}
        \end{tabular}
    }
    \caption{NodePFN enables universal node classification. (a) Each real-world graph requires its own trained GNN model.
    (b) Pre-training on synthetic graphs sampled from controlled priors (varying homophily ($h$) from 0.1 to 0.9) produces \emph{a single model} capable of generalization to arbitrary graphs.} 
    \label{fig:compare}
\end{figure}

We propose a different approach by training on synthetic graphs that systematically cover the diversity of real-world graphs (see \Cref{fig:compare}).
The key insight is learning the posterior predictive distribution (PPD) from synthetic priors. Recently, prior-fitted networks (PFNs) have demonstrated that models trained on synthetic data from carefully designed priors can approximate PPDs for new datasets in a single forward pass~\citep{muller2022transformers,hollmann2023tabpfn}. 
This approach enables in-context learning. That is, the model learns to extract patterns from context examples (labeled nodes) and apply them to query points (unlabeled nodes), enabling immediate prediction without gradient updates.
We extend this PFN paradigm to graphs by designing synthetic graph priors that systematically control homophily levels, community structures, and feature-label relationships. 
We aim to design a model that predicts the label distribution of query nodes based on labeled context nodes in real graphs, by learning PPD from various synthetic graphs.

We introduce \textbf{NodePFN} learning PPDs for node classification from synthetic graph priors. During training, we generate thousands of diverse synthetic graphs, leveraging methods that control class homophily and community levels to ensure that they include a range of network characteristics found in real-world benchmarks.

Experimental evaluation on 23 real-world benchmarks shows that NodePFN achieves competitive performance. Our approach outperforms on both homophily and heterophily graph benchmarks, surpassing GNN baselines. These extensive experiments validate that the patterns governing node classification can \emph{de facto} be learned from synthetic priors.

The contributions of our proposed NodePFN\footnote{Our code is available here: \url{https://github.com/jeongwhanchoi/NodePFN}} are summarized as follows.
\begin{itemize}[leftmargin=*, itemsep=0pt, topsep=2pt]
    \item To the best of our knowledge, we are the first to extend the PFN paradigm to graphs, demonstrating that PPDs for node classification can be learned from synthetic graph prior distributions without requiring actual training data. (\Cref{sec:method}).
    \item  We design a comprehensive synthetic graph prior by using random networks, incorporating levels of homophily, community structure, and feature-label relationships. (\Cref{sec:priors}).
    \item To enable learning graph-aware context from both labeled examples and topological structure, we developed a novel dual-branch architecture combining a context-query attention mechanism with local message passing (\Cref{sec:architecture}).
    \item We demonstrate universal node classification across 23 diverse real-world benchmarks using a single pre-trained model, achieving an average accuracy of 71.27\% and strong performance of 65.14\% on challenging heterophily graphs where traditional GNNs struggle (\Cref{sec:exp}).
\end{itemize}

\section{Preliminaries}\label{sec:pre}
In this section, we introduce posterior predictive distribution and prior-data fitted networks. Then, we address the notation used in our study and node classification

\subsection{Posterior Predictive Distribution in Supervised Learning}
In supervised learning, the goal is to predict labels for unlabeled data points using labeled training samples. 
Given a training set $\mathcal{D}_{\text{train}} = \{(\mathbf{x}_i, \mathbf{y}_i)\}_{i=1}^{n}$ and test set $\mathcal{D}_{\text{test}} = \{\mathbf{x}_j\}_{j=1}^{m}$,  we aim to predict labels for the test set.
In the Bayesian framework, we model the conditional distribution $p(\mathbf{y}|\mathbf{x};\phi)$ with parameters $\phi$ treated as random variables with prior $p(\phi)$. The goal is to predict labels for a test point $\mathbf{x}_{\text{test}}$ using the posterior predictive distribution (PPD):
\begin{align}
p(\mathbf{y}_{\text{test}} | \mathbf{x}_{\text{test}}, \mathcal{D}_{\text{train}}) = \int p(\mathbf{y}_{\text{test}} | \mathbf{x}_{\text{test}}, \phi) p(\phi | \mathcal{D}_{\text{train}})d\phi,\label{eq:ppd}
\end{align}
where the posterior distribution follows Bayes' rule:
\begin{align}
p(\phi|\mathcal{D}_{\text{train}}) \propto p(\phi) \prod_{i=1}^{n} p(\mathbf{y}_i|\mathbf{x}_i; \phi).
\end{align}
If the hypothesis class includes the true conditional distribution, there exists a $\phi^\ast$ such that $p(\mathbf{y} | \mathbf{x};\phi^\ast)=p_{\text{true}}(\mathbf{y} | \mathbf{x})$ for all $(\mathbf{x}, \mathbf{y})$, then the PPD results in optimal prediction.

\subsection{Prior-Data Fitted Networks}\label{sec:pfn}
Prior-data fitted networks (PFNs)~\citep{muller2022transformers} learn an approximation of the PPD from the training data using neural networks. 
Instead of computing the integral in \Cref{eq:ppd} at test time, PFNs are trained on synthetic datasets sampled from a prior $p(\mathcal{D})$ to learn:
\begin{align}
f_\theta: (\mathbf{x}_{\text{test}}, \mathcal{D}_{\text{train}}) \mapsto p(\mathbf{y}_{\text{test}} | \mathbf{x}_{\text{test}}, \mathcal{D}_{\text{train}}).
\end{align}

During training, we sample synthetic datasets from a prior $p(\mathcal{D})$. Each dataset is split into training and test sets. The PFN $f_\theta$ with parameters $\theta$ is trained to minimize the expected loss:
\begin{align}
\mathcal{L}(\theta) = \mathbb{E}_{\mathcal{D} \sim p(\mathcal{D})} \left[ -\log q_\theta(\mathbf{y}_{\text{test}} | \mathbf{x}_{\text{test}}, \mathcal{D}_{\text{train}}) \right],
\end{align}
where $q_\theta$ is the neural network's approximation of the true PPD. By training on synthetic datasets, the model learns to extract relevant patterns from context samples and apply them to new queries.

This approach allows the model to perform inference at test time in a single forward pass without gradient updates, given a new dataset. Through implicit Bayesian inference, the network learns to marginalize over parameter uncertainty. 

\subsection{GNNs for Node Classification and Their Limitations}
In the node classification problem, given a graph $\mathcal{G} = (\mathcal{V}, \mathcal{E})$ with node feature matrix $\mathbf{X} \in \mathbb{R}^{|\mathcal{V}| \times d}$ where $d$ is the feature dimension, adjacency matrix $\mathbf{A} \in \{0, 1\}^{|\mathcal{V}| \times |\mathcal{V}|}$, and a set of labeled nodes $\mathcal{V}_{\text{train}} \subset \mathcal{V}$ with their corresponding labels $\mathbf{y}_{\text{test}}$, the goal is to predict labels $\hat{\mathbf{y}}_{\text{test}}$ for the unlabeled node set $\mathcal{V}_{\text{test}} = \mathcal{V} \setminus \mathcal{V}_{\text{train}}$.

\paragraph{Homophily in Node Classification.}

The success of GNNs is believed to be rooted in the homophily assumption~\citep{mcpherson2001birds}, which implies that connected nodes tend to share similar attributes~\citep{hamilton2020graph}.
This provides additional useful information in the aggregated features compared to the original node features, and the effectiveness of node classification can be determined by the level of edge homophily~\citep{luan2023graph,zhu2020beyond}, which measures the tendency of connected nodes to share the same class label.
The level of homophily, $h$, falls within the range of $[0,1]$, with a value closer to 1, strong homophily, implying that GNNs are more likely to outperform than non-graph models, and vice versa.

\begin{wrapfigure}{r}{0.5\textwidth}
    \centering
    \vspace{-2em}
    \subfigure[]{\includegraphics[width=0.24\linewidth]{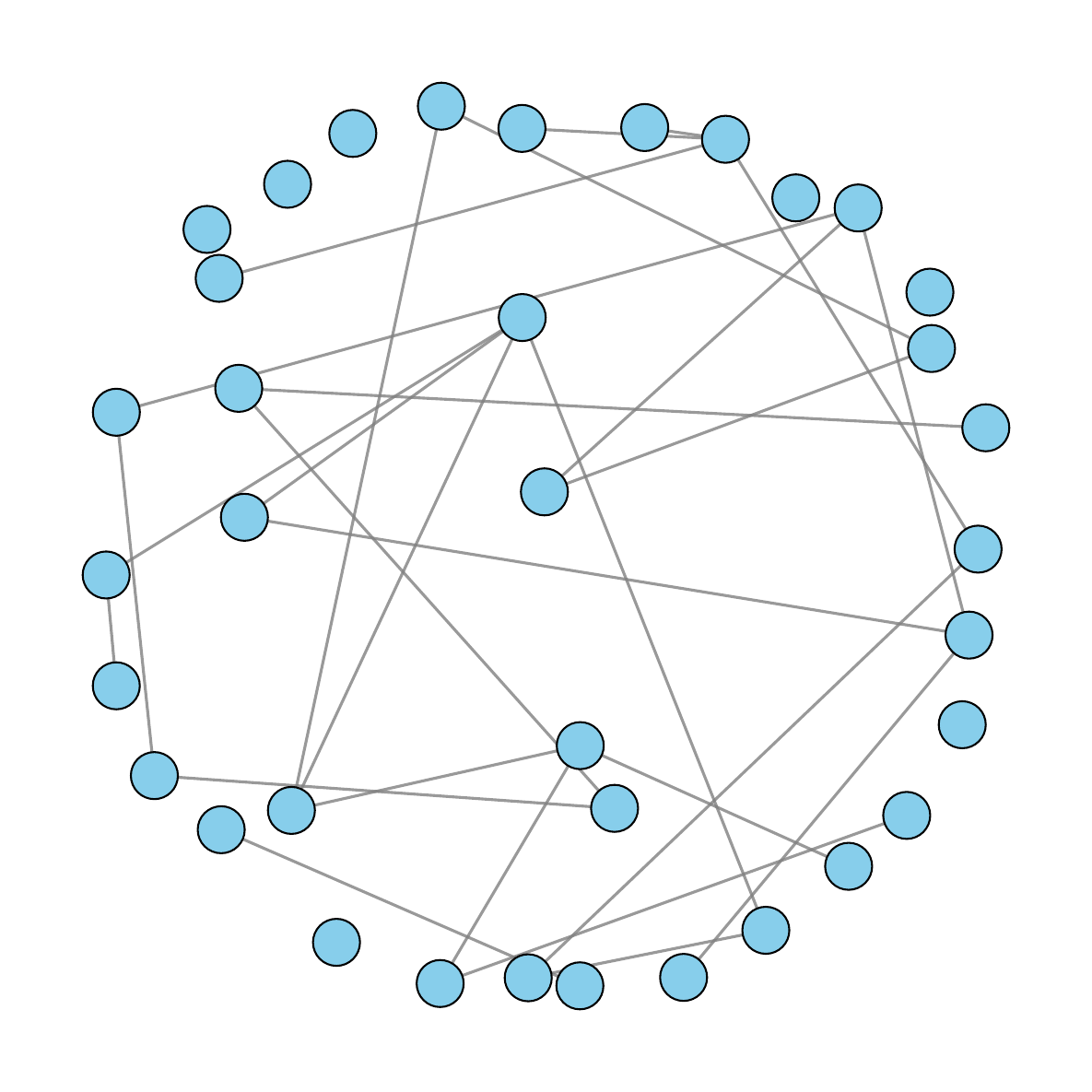}}
    \subfigure[]{\includegraphics[width=0.24\linewidth]{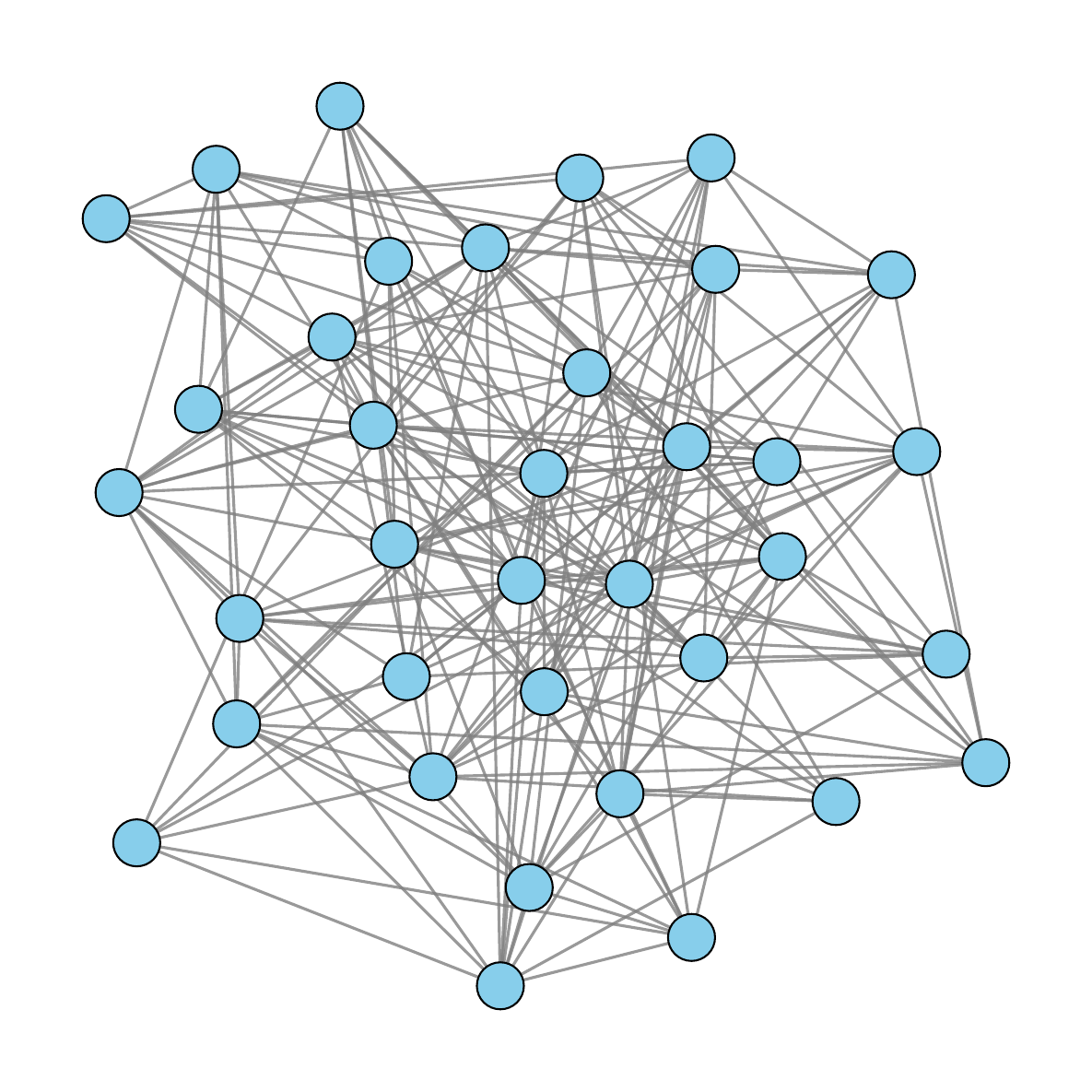}}
    \subfigure[]{\includegraphics[width=0.24\linewidth]{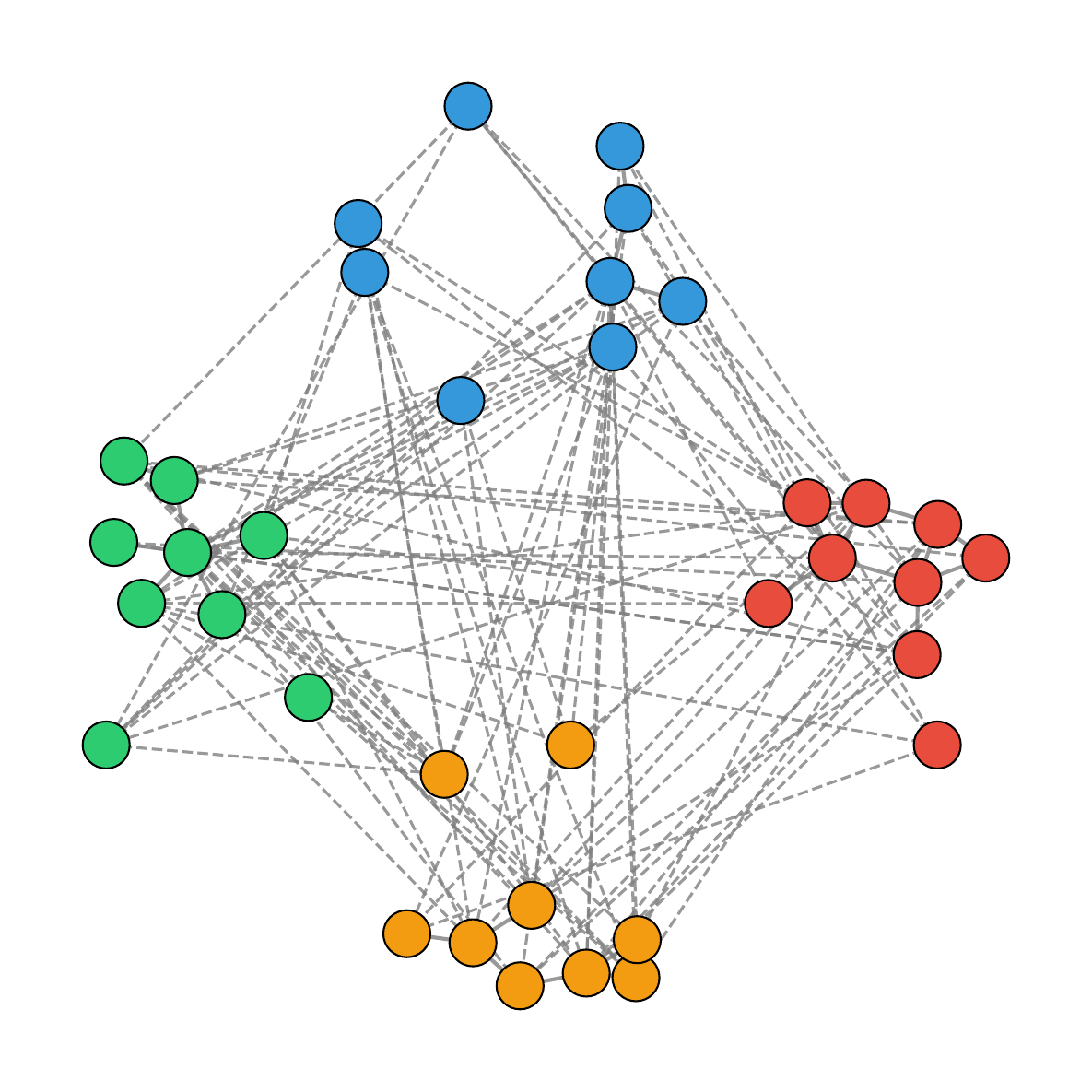}}
    \subfigure[]{\includegraphics[width=0.24\linewidth]{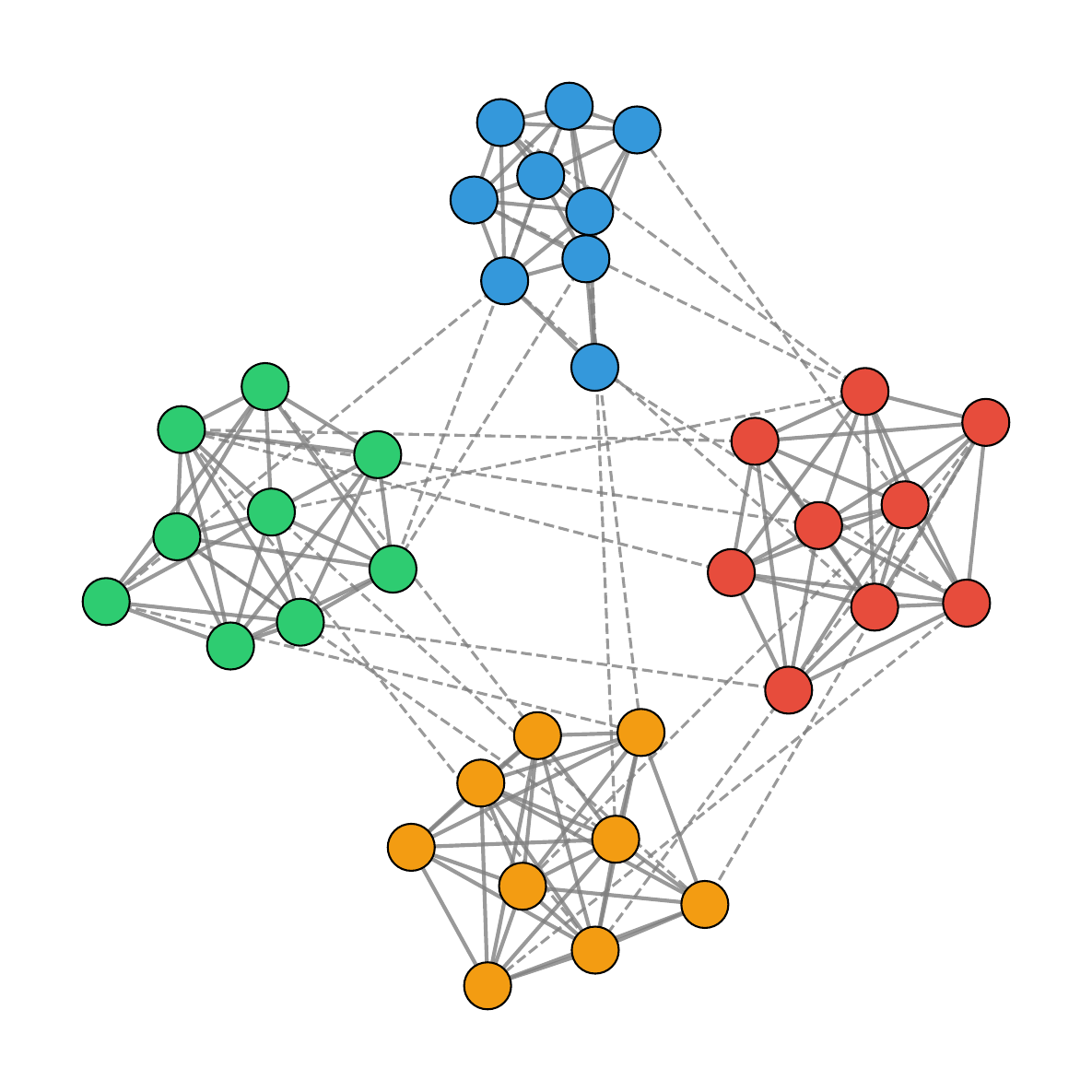}}

    \caption{Network examples with various graph priors used in NodePFN training. (a) Sparse ER. (b) Dense ER. (c) Low homophily cSBM. (d) High homophily cSBM. For simplicity, class color-coded nodes are not shown in ER.}
    \label{fig:vis}
    \vspace{-1em}
\end{wrapfigure}
\subsection{Random Graph Models}
We consider 2 random networks. Erd\H{o}s-Rényi (ER) model~\citep{erdds1959random} generates graphs where each edge appears independently with probability $p_{\mathrm{er}}$. 
This creates graphs with binomial degree distributions and no inherent community structure.
Stochastic block models (SBMs)~\citep{holland1983stochastic} control community structure through different connection probabilities within and between groups.
Contextual SBMs (cSBMs)~\citep{binkiewicz2017covariate} extend SBM by relating community membership to node labels and allow control over homophily.

\section{NodePFN: Prior-Fitted Networks for Node Classification}\label{sec:method}
We introduce NodePFN, a prior-fitted network that learns to approximate PPDs for node classification from synthetic graph data (see \Cref{fig:overview}). Unlike traditional GNNs that require task-specific training, NodePFN performs in-context learning on arbitrary graphs in a single forward pass.

\subsection{Learning from Synthetic Graph Priors}
Given the PPD framework from \Cref{sec:pre}, we train a neural network $f_\theta$ to approximate posterior predictive distributions for node classification. During training, we sample synthetic graphs $\mathcal{G} \sim p(\mathcal{G})$ and learn to predict query node labels from context examples:
\begin{align}
    f_\theta: (\mathbf{x}_{\text{test}}, \mathcal{D}_{\text{train}}, \mathcal{G}) \mapsto p(\mathbf{y}_{\text{test}} | \mathcal{D}_{\text{train}}, \mathcal{G}),
\end{align}
where $\mathcal{D}_{\text{train}} = \{(\mathbf{x}_v, \mathbf{y}_v) : v \in \mathcal{V}_{\text{train}}\}$ contains labeled training nodes. This formulation naturally induces in-context learning: the model learns to extract patterns from training nodes and apply them to test nodes.

\subsection{Synthetic Graph Priors}\label{sec:priors}
\begin{figure}[t]
    \centering
    \resizebox{0.95\linewidth}{!}{%
        \subfigure[Prior Fitting Phase]{\includegraphics[width=0.32\textwidth ,trim={10 12 10 50},clip ]{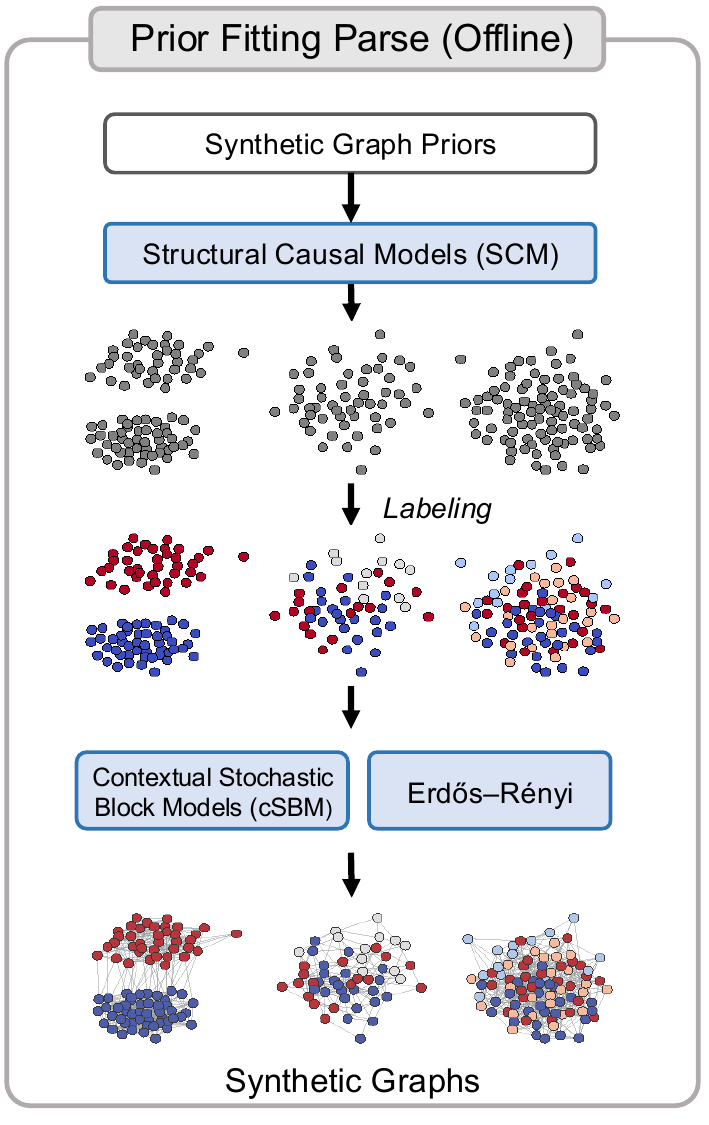}\label{fig:main-a}}
    \hfill
    \subfigure[Pretraining on NodePFN]{\includegraphics[width=0.32\textwidth ,trim={10 12 10 50},clip  ]{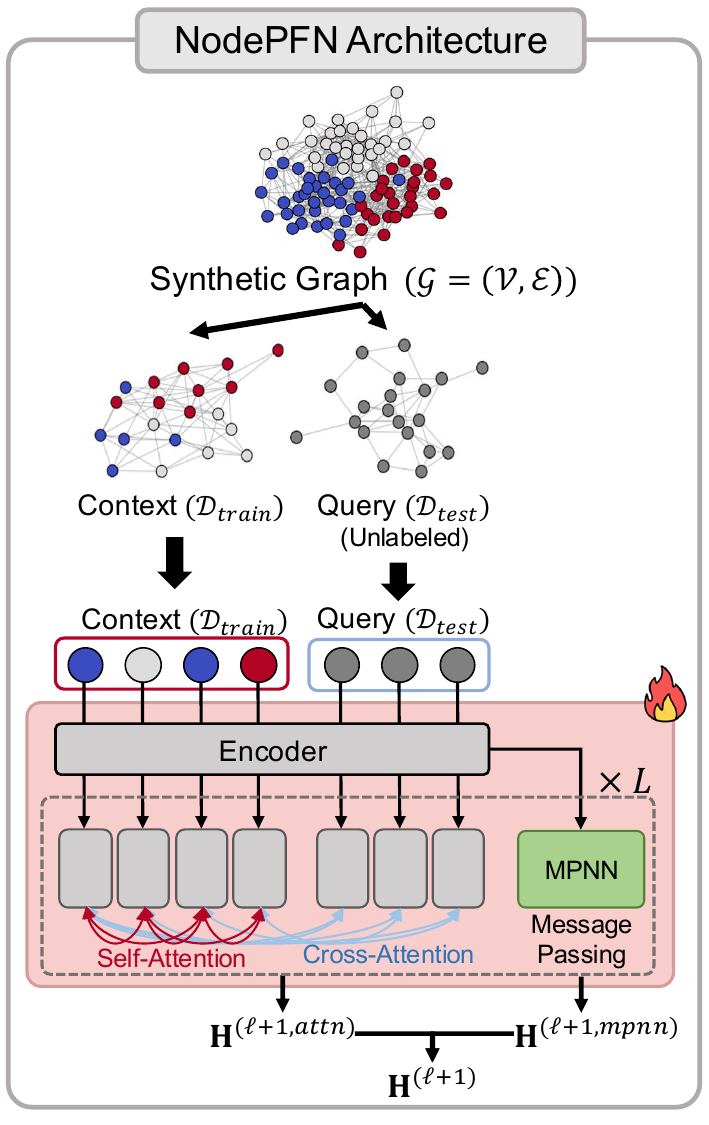}\label{fig:main-b}}
    \hfill
    \subfigure[Inference on Real Graph]{\includegraphics[width=0.32\textwidth ,trim={10 12 10 50},clip  ]{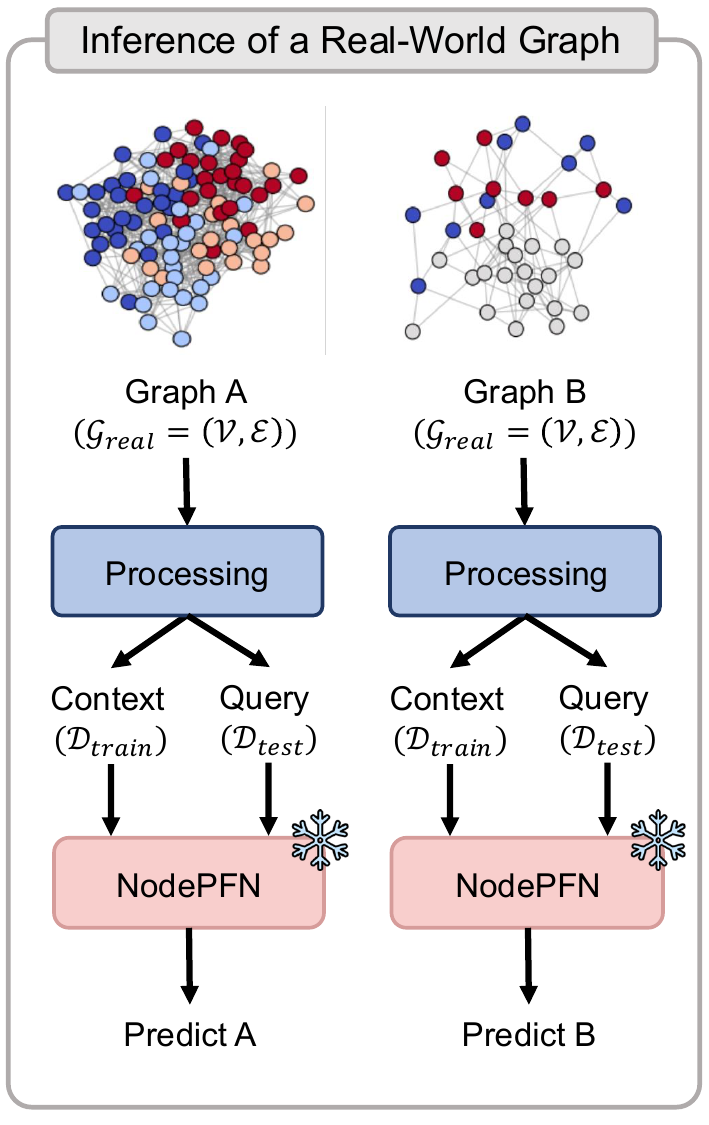}\label{fig:main-c}}}
    \caption{NodePFN overview. (a) Generation of diverse synthetic graph priors with varying structural properties. (b) Dual-branch architecture combining local message passing with context-query attention for in-context learning. (c) Inference on real-world graphs via the pre-trained NodePFN without task-specific re-training. }
    \label{fig:overview}
\end{figure}

As shown in \Cref{fig:main-a}, our approach begins with sampling diverse synthetic graph priors that capture the broad spectrum of structural patterns found in real-world networks. 

\paragraph{Feature-Label Relationships via Causal Models.}
We generate feature-label relationships using structural causal models (SCMs)~\citep{peters2017elements,pearl2009causality} instantiated as random MLPs. For each graph, we sample an MLP architecture and convert it to a DAG by dropping random connections. Gaussian noise propagates through this network to produce node features $\mathbf{X}$ from intermediate layers and labels $y$ from later layers, creating complex non-linear dependencies. 
Importantly, for cSBM graphs, these generated labels determine the community assignments, which in turn control the graph structure through the homophily parameter $h$.

\paragraph{Graph Structure Generation.} 
We use two random network models as shown in \Cref{fig:vis}.
(i) cSBMs generate graphs with controlled community structure and homophily. We sample the homophily level from 0.1 to 0.9. The cSBM creates edges with intra-community probability $p_{\text{in}}$ and inter-community probability $p_{\text{out}}$ such that $h = p_{\text{in}}/(p_{\text{in}} + p_{\text{out}})$. This control over homophily allows us to generate graphs ranging from strong homophily to heterophily.
(ii) ER networks provide unstructured baseline graphs where edges appear independently with probability $p_{\text{er}}$. This ensures the model learns beyond community-based patterns. The distribution for $p_{\text{er}}$ generates graphs with varying densities, from sparse to dense networks.
During training, we sample from both networks to ensure comprehensive coverage of graph structures encountered in practice.

\subsection{Model Architecture}\label{sec:architecture}
Each NodePFN layer consists of two parallel branches that process information complementarily, as shown in~\Cref{fig:main-b}.

\paragraph{Context-Query Attention Branch.}
Following the PFN design~\citep{muller2022transformers}, we use asymmetric attention patterns to enable in-context learning. The initial representations $\mathbf{H}_{\text{train}}^{(0)}$ combine embeddings of both features and labels, while $\mathbf{H}_{\text{test}}^{(0)}$ uses only feature embeddings (detailed implementation in \Cref{app:imp-emb}). For $\mathcal{V}_{\text{train}}$ with observed labels, self-attention allows them to build a comprehensive understanding of the label distribution:
\begin{align}
    \mathbf{H}_{\text{train}}^{(\ell+1,\text{attn})} = \text{SelfAttention}(\mathbf{H}_{\text{train}}^{(\ell)}, \mathbf{H}_{\text{train}}^{(\ell)}, \mathbf{H}_{\text{train}}^{(\ell)}).
\end{align}
For test nodes $\mathcal{V}_{\text{test}}$, cross-attention to training nodes enables leveraging the learned patterns:
\begin{align}
    \mathbf{H}_{\text{test}}^{(\ell+1,\text{attn})} = \text{CrossAttention}(\mathbf{H}_{\text{test}}^{(\ell)}, \mathbf{H}_{\text{train}}^{(\ell)}, \mathbf{H}_{\text{train}}^{(\ell)}),
\end{align}
where the attention functions follow the standard formulation. We employ multiple attention heads with outputs concatenated and linearly projected. This asymmetry ensures test nodes leverage training information without influencing each other's predictions.

\textbf{Local MPNN Branch.}
In parallel, message passing aggregates neighborhood information to capture local graph topology:
\begin{align}
    \mathbf{H}^{(\ell+1,\text{mpnn})} = \mathrm{MPNN}(\mathbf{H}^{(\ell)}, \tilde{\mathbf{A}} ),
\end{align}
where $\tilde{\mathbf{A}} = \mathbf{D}^{-1/2}\mathbf{A}\mathbf{D}^{-1/2}$ is the symmetrically normalized adjacency matrix and $\mathbf{D}$ is a degree matrix. This branch captures structural patterns critical for classification regardless of train/test splits. In our framework, we use GCN~\citep{kipf2017GCN} for the local MPNN branch.

\paragraph{Layer Fusion.}
The parallel branches merge with the input via residual connections:
\begin{align}
    \mathbf{H}^{(\ell+1)} = \text{LayerNorm}(\mathbf{H}^{(\ell)} + \mathbf{H}^{(\ell+1,\text{attn})} + \mathbf{H}^{(\ell+1,\text{mpnn})}).\label{eq:fusion}
\end{align}
This design enables NodePFN to simultaneously learn from labeled examples via attention and local graph structure via message passing. 

\subsection{How to Train}
We train NodePFN to approximate the PPD by minimizing the expected cross-entropy over synthetic graphs sampled from our prior:
\begin{align}
    \mathcal{L}(\theta) = \mathbb{E}_{\mathcal{D} \sim p(\mathcal{D})} \left[ -\frac{1}{|\mathcal{V}_{\text{test}}|} \sum_{v \in \mathcal{V}_{\text{test}}} \sum_{c=1}^{C} y_{v,c} \log f_\theta(y_{v,c} | \mathbf{x}_v, \mathcal{D}_{\text{train}}, \mathcal{G}) \right],
\end{align}
where $C$ is the number of classes, $y_{v,c}$ is the one-hot encoded label for node $v$ and class $c$, 
and $f_\theta$ is our neural approximation to the true PPD from~\Cref{eq:ppd}. 
For each synthetic graph $\mathcal{G}$, we randomly partition nodes into $\mathcal{V}_{\text{train}}$ and $\mathcal{V}_{\text{test}}$.

\subsection{How to Inference}
As shown in ~\Cref{fig:main-c}, NodePFN performs direct prediction on a real-world graph $\mathcal{G}_{\text{real}}$ with its own training-test split. Given labeled nodes $\mathcal{V}_{\text{train}}$ with $\mathcal{D}_{\text{train}} = \{(\mathbf{x}_i, y_i) : i \in \mathcal{V}_{\text{train}}\}$ and unlabeled nodes $\mathcal{V}_{\text{test}}$, the model computes predictions in a single forward pass.

Given a real-world graph, we perform a preprocessing step (\Cref{app:inference}) on the graph and its features.
Then, the model processes the graph through $L$ NodePFN layers and outputs the PPD:
\begin{align}
    f_\theta(y_v | \mathbf{x}_v, \mathcal{D}_{\text{train}}, \mathcal{G}_{\text{real}}) = \text{softmax}(\mathbf{W}_{\text{out}} \mathbf{h}_v^{(L)}),
\end{align}
for each test node $v \in \mathcal{V}_{\text{test}}$. Training nodes incorporate label information through concatenation with features, while test nodes use only features. This provides calibrated uncertainty estimates as the model has learned to approximate the true PPD during training. Importantly, no gradient updates or fine-tuning are required --- the pre-trained model generalizes directly to new graphs.

\section{Experiments}\label{sec:exp}
In this section, we present experiments to evaluate the performance of our proposed NodePFN. We begin by detailing the experimental settings. Next, we investigate the following research questions:
\begin{itemize}[leftmargin=*, itemsep=0pt, topsep=2pt]
    \item \textbf{(RQ1.)} Does our NodePFN perform well on various controlled homophily synthetic graphs?
    \item \textbf{(RQ2.)} Does our NodePFN generalize well for node classification on real-world benchmarks?
    \item \textbf{(RQ3.)} How does the performance of NodePFN compare against training-free methods?
    \item \textbf{(RQ4.)} Does NodePFN perform well compared to the baseline for structural node classification?
    \item \textbf{(RQ5.)} How do components contribute to NodePFN's effectiveness?
\end{itemize}

\begin{wrapfigure}{r}{0.35\textwidth}
    \vspace{-1.5em}
    \includegraphics[width=0.35\textwidth]{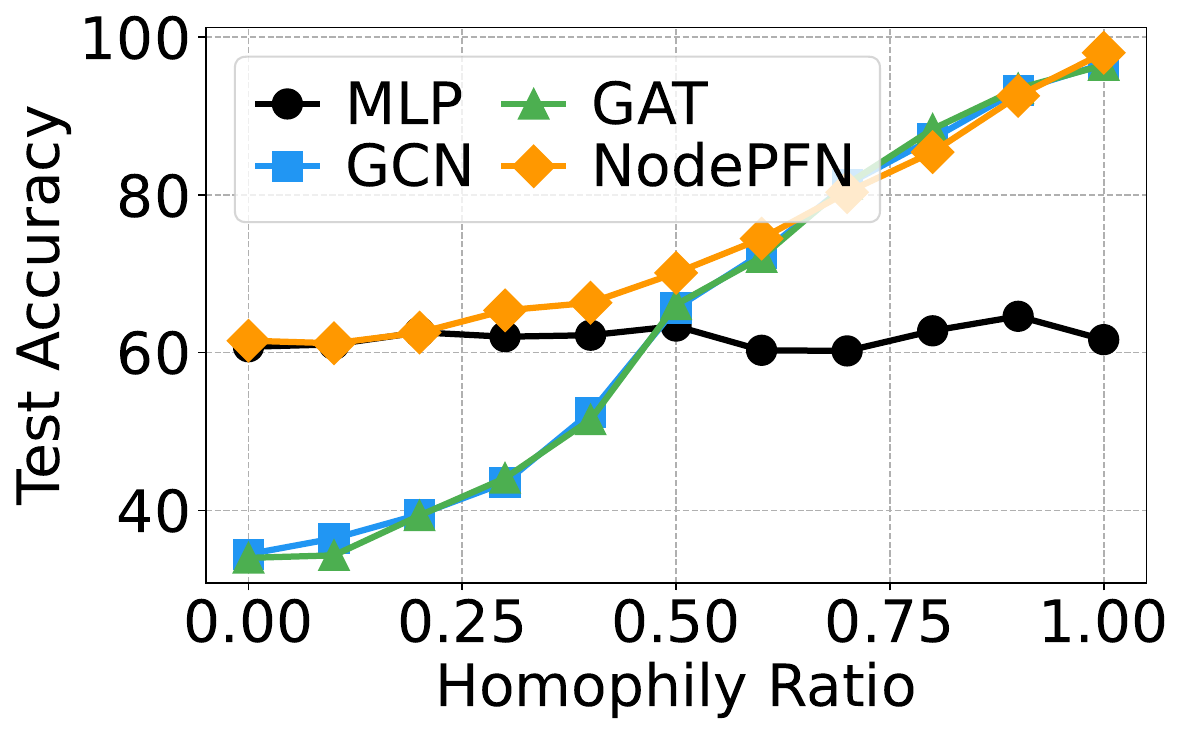}
      \vspace{-1.5em}
    \caption{Experiments on the synthetic Cora Dataset.}
    \label{fig:syn_cora}
    \vspace{-3em}
\end{wrapfigure}

\subsection{\textbf{(RQ1.)} Controlled Synthetic Graphs}
\paragraph{Setup.} To evaluate the classification capability on various homophily ratios, we use the synthetic Cora generator~\cite{li2021deeprobust}. The detailed synthetic datasets are in~\Cref{app:setting}.
\paragraph{Results.}\Cref{fig:syn_cora} shows the mean test accuracy. MLP maintains its test accuracy for all homophily rates. GCN and GAT perform poorly at low homophily rates. Our NodePFN has the best overall trend without sudden drops. Our prior data contribute to its stable accuracy for both homophily and heterophily settings compared with other models.

\begin{table}[ht!]
\centering
    \scriptsize
    \setlength{\tabcolsep}{3pt}
    \renewcommand{\arraystretch}{1.1}
    \caption{Performance comparison on homophily and heterophily real-world benchmark datasets. 
    We report the average accuracy and ranking on each type of dataset, as well as the overall values.}
    \label{tab:results_all}
    \begin{tabular}{c l ccc cccc >{\columncolor[gray]{0.9}}c}

    \toprule
    & Dataset & MLP & GCN & GAT & \makecell{GraphAny\\(Products)} 
    & \makecell{GraphAny\\(Arxiv)} 
    & \makecell{GraphAny\\(Wisconsin)} 
    & \makecell{GraphAny\\(Cora)} & NodePFN \\
    \cmidrule(lr){1-10}
    \multirow{15}{*}{\rotatebox{90}{Homophily Graphs}}
    & AirBrazil    & 23.08\std{5.83} & 42.31\std{7.98} & 57.69\std{14.75} & 34.61\std{16.54} & 34.61\std{16.09} & 36.15\std{16.68} & 33.07\std{16.68} & 75.38\std{1.88} \\
    & AirEU        & 21.25\std{2.31} & 41.88\std{3.60} & 32.50\std{8.45}  & 41.75\std{6.84}  & 41.50\std{6.50}  & 41.13\std{6.02}  & 40.50\std{7.01}  & 57.00\std{1.21}\\
    & AirUS        & 22.88\std{1.46} & 46.49\std{1.81} & 48.47\std{4.17}  & 43.57\std{2.07}  & 43.64\std{1.83}  & 43.86\std{1.44}  & 43.46\std{1.40}  & 61.66\std{0.31}\\
    & Cora         & 48.42\std{0.63} & 81.40\std{0.70} & 81.70\std{1.43}  & 79.36\std{0.23}  & 79.38\std{0.16}  & 77.82\std{1.15}  & 80.18\std{0.13}  & 82.06\std{0.29} \\
    & Citeseer     & 44.40\std{0.44} & 63.40\std{0.63} & 69.10\std{1.59}  & 67.94\std{0.29}  & 68.34\std{0.23}  & 67.50\std{0.44}  & 68.90\std{0.07}  & 67.30\std{0.83} \\
    & Pubmed       & 69.50\std{1.79} & 76.60\std{0.32} & 77.30\std{0.60}  & 76.54\std{0.34}  & 76.36\std{0.17}  & 77.46\std{0.30}  & 76.60\std{0.31}  & 78.00\std{0.24} \\
    & WikiCS       & 72.72\std{0.43} & 79.12\std{0.45} & 79.27\std{0.20}  & 75.01\std{0.54}  & 74.95\std{0.61}  & 73.77\std{0.83}  & 74.39\std{0.71}  & 75.98\std{0.80}\\
    & Amazon-Photo & 68.20\std{0.88} & 91.88\std{0.79} & 91.86\std{1.07}  & 90.64\std{0.82}  & 90.60\std{0.82}  & 90.18\std{0.91}  & 90.14\std{0.93}  & 90.53\std{0.13} \\
    & Amazon-Comp  & 58.28\std{2.98} & 85.83\std{0.86} & 87.01\std{0.50}  & 82.90\std{1.25}  & 83.04\std{1.24}  & 82.00\std{1.14}  & 82.99\std{1.22}  & 81.42\std{0.48} \\
    & DBLP         & 56.27\std{0.62} & 73.02\std{2.22} & 73.87\std{1.35}  & 70.62\std{0.97}  & 70.90\std{0.88}  & 70.13\std{0.77}  & 71.73\std{0.94}  & 74.71\std{0.39} \\
    & Coauthor CS  & 85.88\std{0.93} & 91.83\std{0.71} & 88.47\std{0.79}  & 90.46\std{0.54}  & 90.45\std{0.59}  & 90.85\std{0.63}  & 90.47\std{0.63}  & 91.55\std{0.32} \\
    & Coauthor Physics & 87.43\std{1.98} & 93.93\std{0.37} & 93.01\std{0.89}  & 92.66\std{0.52}  & 92.69\std{0.52}  & 92.54\std{0.43}  & 92.70\std{0.54}  & 93.43\std{0.13} \\
    & Deezer       & 54.24\std{2.15} & 53.69\std{2.29} & 55.99\std{3.78}  & 52.09\std{2.78}  & 52.11\std{2.79}  & 52.13\std{3.02}  & 51.98\std{2.79}  & 53.45\std{0.65} \\
    \cmidrule(lr){2-10}
    \multicolumn{2}{c}{Average Accuracy} & 56.43 & 73.05 & 74.39 & 71.09 & 71.14 & 70.86 & 71.45 & \textbf{77.39} \\
    \multicolumn{2}{c}{Average Ranking} & 7.62 & 4.92 & 4.54 & 4.46 & 4.31 & 4.31 & 4.15 & \textbf{1.69} \\
    \cmidrule(lr){1-10}
    \multirow{12}{*}{\rotatebox{90}{Heterophily Graphs}}
    & Cornell      & 67.57\std{5.06} & 35.14\std{6.51} & 35.14\std{3.52}  & 64.86\std{0.00}  & 65.94\std{1.48}  & 66.49\std{1.48}  & 64.86\std{1.91}  & 71.89\std{2.76} \\
    & Texas        & 48.65\std{4.01} & 51.35\std{2.71} & 54.05\std{2.41}  & 73.52\std{2.96}  & 72.97\std{2.71}  & 73.51\std{1.21}  & 71.89\std{1.48}  & 76.22\std{7.53} \\
    & Wisconsin    & 66.67\std{3.51} & 37.25\std{1.64} & 52.94\std{3.10}  & 65.89\std{2.23}  & 65.10\std{3.22}  & 71.77\std{5.98}  & 61.18\std{5.08}  & 79.22\std{6.97} \\
    & Chameleon    & 38.87\std{2.21} & 41.31\std{3.05} & 39.83\std{2.10}  & 39.45\std{4.20}  & 37.40\std{3.11}  & 36.67\std{5.32}  & 37.99\std{4.54}  & 50.13\std{3.30} \\
    & Actor        & 33.95\std{0.80} & 28.55\std{0.68} & 27.30\std{0.22}  & 28.99\std{0.61}  & 28.60\std{0.21}  & 29.51\std{0.55}  & 27.91\std{0.16}  & 32.99\std{1.09} \\
    & Minesweeper  & 80.00\std{0.00} & 81.12\std{0.37} & 80.08\std{0.04}  & 80.27\std{0.16}  & 80.30\std{0.13}  & 80.13\std{0.09}  & 80.46\std{0.15}  & 80.66\std{0.25} \\
    & Tolokers     & 78.16\std{0.02} & 79.93\std{0.10} & 78.50\std{0.55}  & 78.18\std{0.03}  & 78.18\std{0.04}  & 78.24\std{0.03}  & 78.20\std{0.02}  & 78.61\std{0.06} \\
    & Amazon-Ratings   & 47.90\std{0.45} & 47.35\std{0.26} & 47.18\std{0.42}  & 42.70\std{0.10}  & 42.74\std{0.12}  & 42.57\std{0.34}  & 42.84\std{0.04}  & 44.68\std{0.48} \\
    & Questions    & 97.33\std{0.06} & 97.15\std{0.04} & 97.11\std{0.02}  & 97.10\std{0.01}  & 97.09\std{0.02}  & 97.11\std{0.00}  & 97.06\std{0.03}  & 97.02\std{0.01} \\
    & Squirrel     & 35.55\std{0.98} & 38.67\std{1.84} & 38.78\std{2.39}  & 38.92\std{2.98}  & 37.73\std{2.31}  & 36.76\std{3.55}  & 37.25\std{2.65}  & 43.40\std{1.03} \\
    \cmidrule(lr){2-10}
    \multicolumn{2}{c}{Average Accuracy} & 58.17 & 58.84 & 59.11 & 61.39 & 60.71 & 61.62 & 60.56 & \textbf{65.14} \\
    \multicolumn{2}{c}{Average Ranking} & 7.20 & 6.80 & 6.60 & 4.90 & 4.70 & 4.50 & 4.60 & \textbf{1.70} \\
  \cmidrule(r){1-10}
    \multicolumn{2}{l}{Avg. Accuracy}& 57.30 & 66.63 & 67.67 & 66.24 & 65.93 & 66.24 & 66.00 & \textbf{71.27} \\ 
     \multicolumn{2}{l}{Avg. Ranking}  & 7.41 & 5.86 & 5.57 & 4.68 & 4.50 & 4.40 & 4.38 & \textbf{1.70} \\ 
     \bottomrule
    \end{tabular}
\end{table}

\subsection{\textbf{(RQ2.)} Experiments on Real-world Graph Benchmarks}
\paragraph{Setup.}
We evaluate on 23 benchmark datasets for node classification.
We compare against MLP, GCN~\citep{kipf2017GCN}, 
GAT~\citep{velickovic2018GAT}, and GraphAny~\citep{zhao2025graphany} models.
If there are reported results from \citet{zhao2025graphany}, we directly adopt the reported results, otherwise, we run experiments with their optimal setting.
More detailed dataset and evaluation settings are provided in~\Cref{app:dataset,app:hyp_nc}.
\paragraph{Results.}
\Cref{tab:results_all} presents a comprehensive comparison of our results on 23 datasets. The results show that NodePFN achieves the best overall average accuracy of 71.27\% using \emph{only a single pre-trained model}. In contrast, GraphAny models require training on each specific dataset but still underperform NodePFN.
On homophily and heterophily datasets, NodePFN achieves the highest average accuracy.
Moreover, GraphAny models show inconsistent performance depending on the characteristics of the training dataset. GraphAny (Cora) performs well on homophily graphs but worse on Wisconsin, one of the heterogeneous graphs.
In contrast, NodePFN consistently performs well on both graph types without requiring dataset-specific training.

\begin{wraptable}{r}{0.45\textwidth}
    \scriptsize
    \centering
    \vspace{-2em} 
    \setlength{\tabcolsep}{1.5pt}
    \caption{Training-free models vs NodePFN.}
    \renewcommand{\arraystretch}{0.8}
    \vspace{-0.7em}
    \label{tab:closedform}
    \begin{tabular}{lcccc}
    \toprule
    Method & Cora & Pubmed & Wisconsin  & Texas \\
    \midrule
     Linear & 52.80\std{0.00} & 59.30\std{0.00} & {80.00\std{2.15}} & 32.35\std{5.30} \\
     SGC   & {78.20\std{0.00}} & {72.98\std{0.00}} & 57.64\std{1.07} & 46.03\std{6.86} \\
     HGC   & 22.50\std{0.00} & 46.32\std{0.00} & 64.32\std{2.51} & {57.54\std{6.30}} \\
     LabelProp          & 60.30\std{0.00} & 63.44\std{0.04} & 16.08\std{2.15} & 23.53\std{5.51} \\
     TFGNN          & 60.03\std{0.00} & 40.04\std{0.01} & 14.51\std{3.01} & 19.91\std{6.10} \\
    \midrule
     \rowcolor{gray!20} NodePFN & {82.06\std{0.29}} & {78.00\std{0.24}} & {81.18\std{5.70}} & {76.22\std{7.53}} \\
    \bottomrule
       \vspace{-3.5em}
    \end{tabular}
\end{wraptable}

\subsection{\textbf{(RQ3.)} Comparison with Training-free Methods}\label{sec:rq3}
\paragraph{Setup.}
NodePFN can be compared with several training-free methods, which can be tested directly without training steps.
We compare closed-form solution methods that use pseudo-inverse operations to solve node classification as a regression problem~\citep{zhao2025graphany}. The methods include the ``Linear'' model that predicts directly without graph convolutions, and closed-form models using SGC~\citep{wu2019simplifying} and high-pass filter graph convolutions (HGC)~\citep{chien2021adaptive,luan2022revisiting}.
We also include label propagation (``LabelProp'')~\citep{zhu2002learning} and TF-GNNs~\citep{sato_training-free_2024}.

\paragraph{Results.}
In \Cref{tab:closedform}, the results demonstrate NodePFN's superior performance on all datasets compared to training-free baselines. 
Our NodePFN consistently outperforms all baseline methods while using a single pre-trained model. This demonstrates that the learned inductive bias of NodePFN surpasses analytical closed-form solutions and highlights the value of the pre-trained approach for generalizable node classification.

\begin{figure}[t]
    \centering
    \begin{minipage}[t]{0.32\textwidth} 
        \centering
        \includegraphics[width=\linewidth]{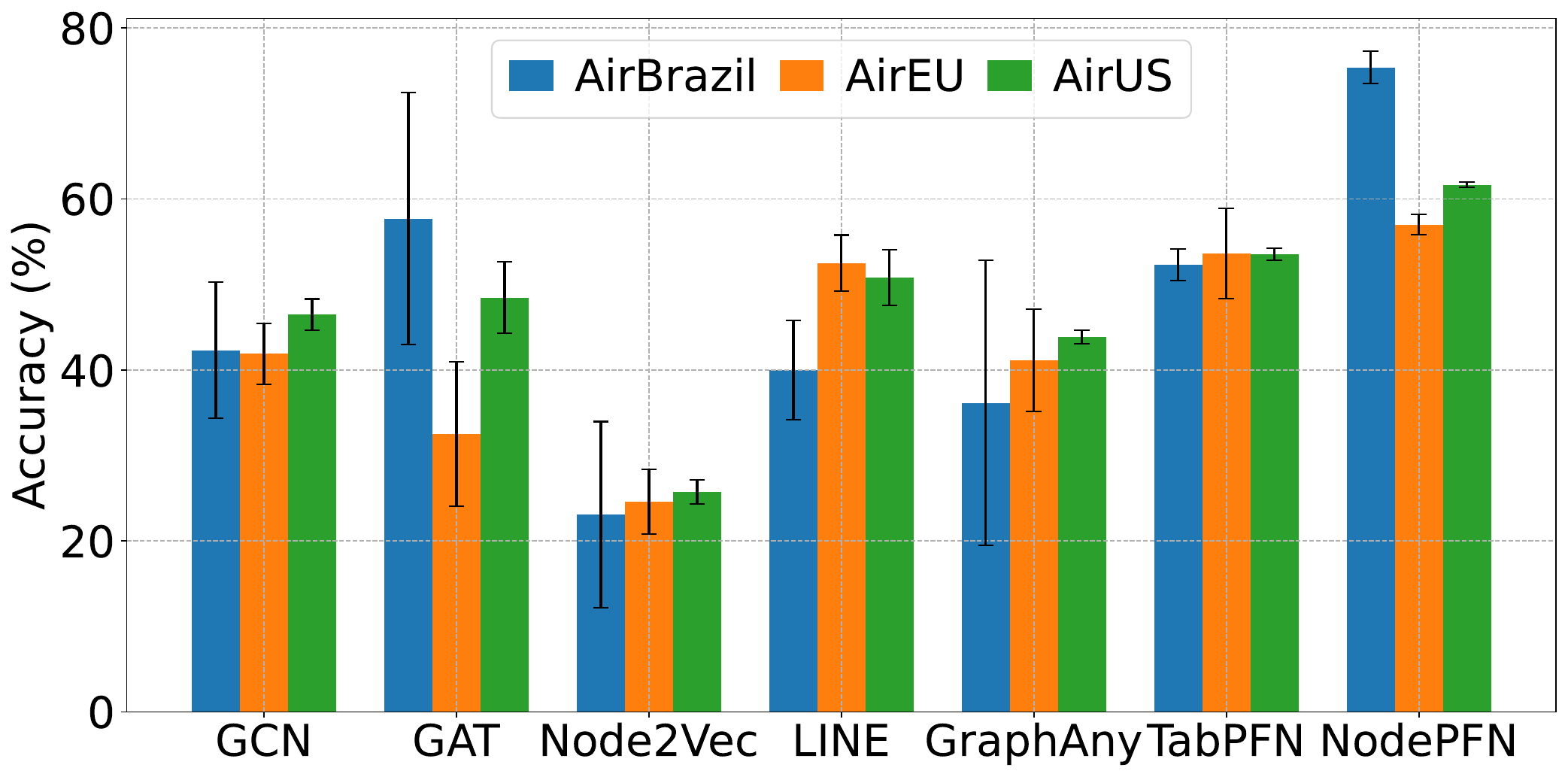}
        \caption{Structural node classification results (GraphAny trained on Wisconsin).}
        \label{fig:airports}
    \end{minipage}
    \hfill
    \begin{minipage}[t]{0.32\textwidth}
        \centering
        \includegraphics[width=\linewidth]{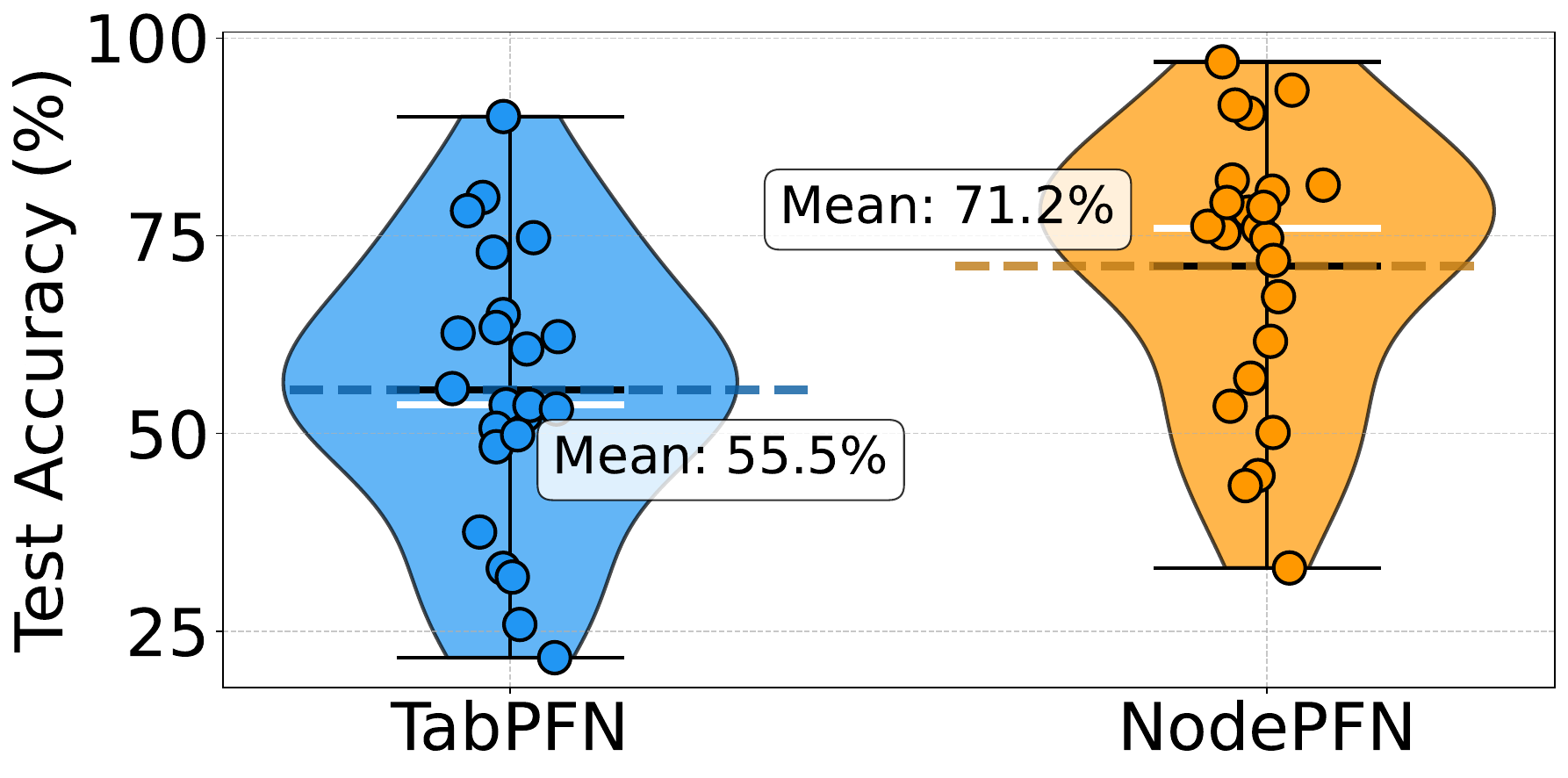}
        \caption{Comparison of accuracy distributions between TabPFN and NodePFN.}
        \label{fig:violin}
    \end{minipage}
    \hfill
    \begin{minipage}{0.32\textwidth}
        \centering
        \scriptsize
        \setlength{\tabcolsep}{2pt}
      \vspace{-2em}
        \captionof{table}{Ablation studies on NodePFN components.}
        \renewcommand{\arraystretch}{0.9}
        \vspace{-1em}
        \label{tab:ablation}
        \begin{tabular}{l ccc}
            \toprule
            Ablation & Cora & Wisconsin & Tolokers\\
            \midrule
            w/o ER & 81.26 & 78.82 & 77.30 \\
            w/o cSBM & 80.62 & 80.39 & 77.18 \\
            \cmidrule(lr){1-4}
            TabPFN & 53.10 & 72.94 & 78.18 \\
            \cmidrule(lr){1-4}
            NodePFN-L6 & 53.10 & 72.94 & 78.00 \\
            NodePFN-Seq & 80.64 & 78.82 & 77.88 \\
            \cmidrule(lr){1-4}
            \rowcolor{gray!20} NodePFN & 82.06 & 81.18 & 78.61 \\
            \bottomrule
        \end{tabular}
    \end{minipage}
\end{figure}

\subsection{\textbf{(RQ4.)} Structural Node Classification}
\paragraph{Setup.}
We evaluate NodePFN on Airport datasets~\citep{ribeiro2017struc2vec}, where the goal is to predict the ``structural role'' of each node based only on the network topology without node features~\citep{cui2022positional}. 
Node features are provided as one-hot encoded node identifiers.
This setting evaluates whether NodePFN can learn structural roles when forced to rely primarily on topological patterns.
We include Node2Vec~\citep{grover2016node2vec} and LINE~\citep{tang2015line} as additional baselines, since these methods specialize in structural embedding.
\paragraph{Results.}
As shown in \Cref{fig:airports}, the results show that NodePFN outperforms all baselines. 
This suggests that NodePFN learns robust structural patterns that generalize beyond node features and effectively identifies meaningful node properties as well as structural roles based on network topology.

\subsection{\textbf{(RQ.5)} Ablation Studies}\label{sec:ablation}
\Cref{tab:ablation} demonstrates the robustness of NodePFN's design through ablation studies. Removing ER Networks or cSBM shows minimal performance degradation: for homophily datasets, Cora, cSBM removal causes slight drops, while for heterophily datasets such as Wisconsin and Tolokers (with 10 features and ~0.5 homophily ratio), the impact is negligible. This indicates that these priors adapt well to different graph characteristics. The architectural variant NodePFN-L6, with reduced model capacity from 29.01M to 14.80M parameters, shows performance drops on Cora. This suggests that sufficient model capacity is important for learning patterns in highly homophily datasets. NodePFN-Seq with sequential processing maintains competitive performance, validating the effectiveness of both parallel and sequential architectures for combining structural information. 

We also compare TabPFN, since when all graph-specific priors and MPNN components are removed, our proposed NodePFN can be reduced to TabPFN. As shown in~\Cref{fig:violin}, NodePFN outperforms TabPFN on all datasets (see \Cref{app:vs_tabpfn} for full results). At the same time, TabPFN shows wider variance and lower overall accuracy, confirming the necessity of graph-aware modeling over treating nodes as independent tabular data. 

\section{Related Work}\label{sec:related}
\paragraph{Prior-data Fitted Networks.}
\citet{muller2022transformers} introduced PFNs and proved that a Transformer trained on tasks drawn from a prior can approximate PPDs from in-context examples. 
Following this work, \citet{nagler2023statistical} shows how PFNs approximate PPD and why they can still learn at inference, and this paradigm has been adapted to specialized domains. TabPFN~\citep{hollmann2023tabpfn,hollmann2025tabpfnv2} demonstrates that carefully designed synthetic priors can yield state-of-the-art performance on small tabular datasets. 
Also, the PFN has been adapted to time-series forecasting~\citep{dooley2023forecastpfn} and \citet{hoo2024tabular,hoo2025tables} analyzes time series via feature engineering and encodes temporal patterns as tabular features. 
Concurrently, TabPFN-GN~\citep{choi2025tabpfngn} converts graph data into tabular features for direct use with TabPFN, but its reliance on feature engineering limits performance on heterophily graphs (cf.\ \Cref{sec:ablation}).

\paragraph{Graph Foundation Models.}
Recent works primarily leverage LLMs for zero-shot learning. 
GraphGPT~\citep{tang2024graphgpt}, GraphLLM~\citep{chai2023graphllm}, and LLAGA~\citep{chen2024llaga} convert graphs to text descriptions, while frameworks that use text-attributed graph datasets~\citep{li2024glbench}, such as OFA~\citep{liu2024ofa}, use LLMs to encode node features. 
More recently, GOFA~\citep{kong2024gofa}, Graph-R1~\citep{wu2025graph}, and ZeroG~\citep{li2024zerog} extend this line of work by exploring joint graph–language modeling, explicit reasoning for zero-shot learning, and cross-dataset transferability, respectively. These approaches leverage LLMs' strengths and limitations, including their dependency on textual attributes. In contrast, our approach requires no LLMs and works with arbitrary node features.

\paragraph{GNNs for Node Classification.}
Node classification is a classical graph machine learning task on which GNNs have recently achieved strong results.
GCN~\citep{kipf2017GCN}, GraphSAGE~\citep{hamilton2017graphSAGE}, and GAT~\citep{velickovic2018GAT} established the foundation of GNNs by showing strong performance on homophily graph datasets. 
Additionally, neighborhood aggregation of GNNs shows stable performance on homophily graph benchmark datasets but struggles with heterophily graphs~\citep{pei2020geom}. 
\citet{sato_training-free_2024} proposes training-free GNNs for node classification, but they are always suboptimal for GNN performance and have limitations that make them inapplicable to heterophily graphs (see \Cref{sec:rq3}).
As GNNs may not dominate all graph networks, zero-shot approaches leveraging pretrained models such as TabPFN can bypass this architecture search.

\section{Discussion}\label{sec:discussion}
Although our primary objective is to demonstrate that universal node classification can be achieved via synthetic graph priors, the proposed NodePFN has limitations. 
NodePFN currently requires fixed maximum class numbers (tested up to 20 classes) and feature dimensions during training, and the attention mechanism's quadratic complexity restricts applicability to large-scale graphs. We leave these limitations for future work.

Despite these limitations, NodePFN's pre-training paradigm offers significant advantages. Although the model requires computational resources to pre-train on approximately 250,000 synthetic graphs (see \Cref{app:prior_scale}), this investment is amortized across all subsequent inference tasks.
This contrasts with conventional GNNs that require retraining for each new dataset. Extensive experiments demonstrate that NodePFN achieves universal node classification, thereby justifying this initial computational overhead.

This universal applicability stems from our focus on learning structural patterns from synthetic priors. Unlike recent graph foundation models that rely on text-attributed graphs and LLMs~\citep{tang2024graphgpt,chai2023graphllm,liu2024ofa} (as discussed in \Cref{sec:related}), NodePFN operates on graphs with arbitrary numerical features without requiring semantic understanding.

\section{Concluding Remarks}
We presented NodePFN, the first extension of PFNs to graphs, showing that universal node classification can be learned from synthetic graph priors. A single NodePFN model demonstrates an average accuracy of 71.27\% on 23 benchmarks, particularly outperforming standard GNNs on heterophily graphs.
This work represents a new paradigm for graph machine learning through synthetic pre-training, validating that universal patterns can be learned without real-world training data.

We discussed the limitations of NodePFN in \Cref{sec:discussion}. Future work could address these limitations by exploring efficient attention mechanisms for massive graphs and investigating hybrid approaches that combine our structural pattern learning with semantic processing from text-attributed graphs. 
\clearpage
\section*{Acknwoledgements}
This work was partly supported by the Institute for Information \& Communications Technology Planning \& Evaluation (IITP) grants funded by the Korean government (MSIT) (No. RS-2024-00457882, AI Research Hub Project; No. RS-2025-25442149, LG AI STAR Talent Development Program for Leading Large-Scale Generative AI Models in the Physical AI Domain), and the Korea Advanced Institute of Science and
Technology (KAIST) grant funded by the Korea government (MSIT)
(No. G04240001, Physics-inspired Deep Learning).
\section*{Ethical Statements}
In terms of the broader impact of this research on society, we do not see the very negative impacts that might be expected.

\section*{Use of LLMs}
In accordance with ICLR 2026 policy, we acknowledge the use of LLMs in the preparation of this paper.
To achieve better grammar, expression, and translation, we use Google Translate~\citep{gtranslate} and DeepL~\citep{deepl} to improve some texts. DeepL has an LLM-powered feature built in.

\section*{Reproducibility Statement}
To ensure reproducibility and completeness, we have included appendices in this paper.
We also report the model architecture, all training hyperparameters, and the hardware specifications for our experiments in \Cref{app:setting}. 
The synthetic graph prior generation and all associated hyperparameters are described in \Cref{app:prior,app:hyp_nc}.
The source code can be found in the following: \url{https://github.com/jeongwhanchoi/NodePFN}
\bibliography{reference}

@String{Springer = "Springer-Verlag" }

@inproceedings{Rex2018pinsage,
  author    = {Rex Ying and
               Ruining He and
               Kaifeng Chen and
               Pong Eksombatchai and
               William L. Hamilton and
               Jure Leskovec},
  title     = {Graph Convolutional Neural Networks for Web-Scale Recommender Systems},
  booktitle = {KDD},
  year      = {2018}
}

@inproceedings{He20LightGCN,
author = {He, Xiangnan and Deng, Kuan and Wang, Xiang and Li, Yan and Zhang, YongDong and Wang, Meng},
title = {LightGCN: Simplifying and Powering Graph Convolution Network for Recommendation},
year = {2020},
booktitle = {SIGIR}
}

@inproceedings{grover2016node2vec,
author = {Grover, Aditya and Leskovec, Jure},
title = {Node2vec: Scalable Feature Learning for Networks},
year = {2016},
booktitle = {KDD},
pages = {855–864},
numpages = {10}
}

@inproceedings{tang2015line,
author = {Tang, Jian and Qu, Meng and Wang, Mingzhe and Zhang, Ming and Yan, Jun and Mei, Qiaozhu},
title = {LINE: Large-Scale Information Network Embedding},
year = {2015},
booktitle = {TheWebConf (former WWW)},
pages = {1067–1077},
numpages = {11}
}

@inproceedings{kipf2017GCN,
  title={Semi-Supervised Classification with Graph Convolutional Networks},
  author={Kipf, Thomas N. and Welling, Max},
  booktitle={International Conference on Learning Representations},
  year={2017}
}

@inproceedings{Gilmer2017chemi,
  author    = {Justin Gilmer and Samuel S. Schoenholz and Patrick F. Riley and Oriol Vinyals and George E. Dahl},
  title     = {Neural Message Passing for Quantum Chemistry},
  booktitle = {International Conference on Machine Learning},
  year      = {2017}
}

@article{bresson2017GatedGCN,
  title={Residual gated graph convnets},
  author={Bresson, Xavier and Laurent, Thomas},
  journal={arXiv preprint arXiv:1711.07553},
  year={2017}
}

@inproceedings{hamilton2017graphSAGE,
  title={Inductive representation learning on large graphs},
  author={Hamilton, Will and Ying, Zhitao and Leskovec, Jure},
  booktitle={Advances in Neural Information Processing Systems},
  year={2017}
}

@inproceedings{velickovic2018GAT,
  title="{Graph Attention Networks}",
  author={Veli{\v{c}}kovi{\'{c}}, Petar and Cucurull, Guillem and Casanova, Arantxa and Romero, Adriana and Li{\`{o}}, Pietro and Bengio, Yoshua},
  booktitle={Proceedings of the International Conference on Learning Representations (ICLR)},
  year={2018}
}

@inproceedings{Klicpera2019APPNP,
  author    = {Johannes Klicpera and Aleksandar Bojchevski and Stephan G{\"{u}}nnemann},
  title     = {Predict then Propagate: Graph Neural Networks meet Personalized PageRank},
  booktitle = {Proceedings of the International Conference on Learning Representations (ICLR)},
  year      = {2019}
}

@inproceedings{Bo2021fagcn,
  title={Beyond Low-frequency Information in Graph Convolutional Networks},
  author={Deyu Bo and Xiao Wang and Chuan Shi and Huawei Shen},
  booktitle = {AAAI},
  year      = {2021}
}

@inproceedings{chien2021GPRGNN,
  title={Adaptive Universal Generalized {P}age{R}ank Graph Neural Network},
  author={Eli Chien and Jianhao Peng and Pan Li and Olgica Milenkovic},
  booktitle={Proceedings of the International Conference on Learning Representations (ICLR)},
  year={2021}
}

@inproceedings{choi2023gread,
  title={GREAD: Graph neural reaction-diffusion networks},
  author={Choi, Jeongwhan and Hong, Seoyoung and Park, Noseong and Cho, Sung-Bae},
  booktitle={International Conference on Machine Learning (ICML)},
  pages={5722--5747},
  year={2023},
  organization={PMLR}
}

@article{hamilton2020graph,
author={Hamilton, William L.},
title={Graph Representation Learning},
journal={Synthesis Lectures on Artificial Intelligence and Machine Learning},
volume={14},
number={3},
pages={1-159},
year = {2020},
publisher={Morgan and Claypool}
}

@article{zhou2020graph,
  title={Graph neural networks: A review of methods and applications},
  author={Zhou, Jie and Cui, Ganqu and Hu, Shengding and Zhang, Zhengyan and Yang, Cheng and Liu, Zhiyuan and Wang, Lifeng and Li, Changcheng and Sun, Maosong},
  journal={AI open},
  volume={1},
  pages={57--81},
  year={2020},
  publisher={Elsevier}
}

@article{achiam2023gpt,
  title={Gpt-4 technical report},
  author={Achiam, Josh and Adler, Steven and Agarwal, Sandhini and Ahmad, Lama and Akkaya, Ilge and Aleman, Florencia Leoni and Almeida, Diogo and Altenschmidt, Janko and Altman, Sam and Anadkat, Shyamal and others},
  journal={arXiv preprint arXiv:2303.08774},
  year={2023}
}

@article{touvron2023llama,
  title={Llama: Open and efficient foundation language models},
  author={Touvron, Hugo and Lavril, Thibaut and Izacard, Gautier and Martinet, Xavier and Lachaux, Marie-Anne and Lacroix, Timoth{\'e}e and Rozi{\`e}re, Baptiste and Goyal, Naman and Hambro, Eric and Azhar, Faisal and others},
  journal={arXiv preprint arXiv:2302.13971},
  year={2023}
}

@article{li2024glbench,
  title={Glbench: A comprehensive benchmark for graph with large language models},
  author={Li, Yuhan and Wang, Peisong and Zhu, Xiao and Chen, Aochuan and Jiang, Haiyun and Cai, Deng and Chan, Victor W and Li, Jia},
  journal={Advances in Neural Information Processing Systems},
  volume={37},
  pages={42349--42368},
  year={2024}
}

@inproceedings{hollmann2023tabpfn,
    title={Tab{PFN}: A Transformer That Solves Small Tabular Classification Problems in a Second},
    author={Noah Hollmann and Samuel M{\"u}ller and Katharina Eggensperger and Frank Hutter},
    booktitle={The Eleventh International Conference on Learning Representations },
    year={2023},
    url={https://openreview.net/forum?id=cp5PvcI6w8_}
}

@article{hollmann2025tabpfnv2,
  title={Accurate predictions on small data with a tabular foundation model},
  author={Hollmann, Noah and M{\"u}ller, Samuel and Purucker, Lennart and Krishnakumar, Arjun and K{\"o}rfer, Max and Hoo, Shi Bin and Schirrmeister, Robin Tibor and Hutter, Frank},
  journal={Nature},
  volume={637},
  number={8045},
  pages={319--326},
  year={2025},
  publisher={Nature Publishing Group UK London}
}

@article{hoo2025tables,
  title={From Tables to Time: How TabPFN-v2 Outperforms Specialized Time Series Forecasting Models},
  author={Hoo, Shi Bin and M{\"u}ller, Samuel and Salinas, David and Hutter, Frank},
  journal={arXiv preprint arXiv:2501.02945},
  year={2025}
}

@inproceedings{hoo2024tabular,
  title={The tabular foundation model tabpfn outperforms specialized time series forecasting models based on simple features},
  author={Hoo, Shi Bin and M{\"u}ller, Samuel and Salinas, David and Hutter, Frank},
  booktitle={NeurIPS Workshop on Time Series in the Age of Large Models},
  year={2024}
}

@inproceedings{zhao2025graphany,
title={Fully-inductive Node Classification on Arbitrary Graphs},
author={Jianan Zhao and Zhaocheng Zhu and Mikhail Galkin and Hesham Mostafa and Michael M. Bronstein and Jian Tang},
booktitle={The Thirteenth International Conference on Learning Representations},
year={2025},
url={https://openreview.net/forum?id=1Qpt43cqhg}
}

@inproceedings{tang2024graphgpt,
  title={Graphgpt: Graph instruction tuning for large language models},
  author={Tang, Jiabin and Yang, Yuhao and Wei, Wei and Shi, Lei and Su, Lixin and Cheng, Suqi and Yin, Dawei and Huang, Chao},
  booktitle={Proceedings of the 47th International ACM SIGIR Conference on Research and Development in Information Retrieval},
  pages={491--500},
  year={2024}
}

@inproceedings{liu2024ofa,
title={One For All: Towards Training One Graph Model For All Classification Tasks},
author={Hao Liu and Jiarui Feng and Lecheng Kong and Ningyue Liang and Dacheng Tao and Yixin Chen and Muhan Zhang},
booktitle={The Twelfth International Conference on Learning Representations},
year={2024},
url={https://openreview.net/forum?id=4IT2pgc9v6}
}

@article{chai2023graphllm,
  title={Graphllm: Boosting graph reasoning ability of large language model},
  author={Chai, Ziwei and Zhang, Tianjie and Wu, Liang and Han, Kaiqiao and Hu, Xiaohai and Huang, Xuanwen and Yang, Yang},
  journal={arXiv preprint arXiv:2310.05845},
  year={2023}
}

@article{brown2020language,
  title={Language models are few-shot learners},
  author={Brown, Tom and Mann, Benjamin and Ryder, Nick and Subbiah, Melanie and Kaplan, Jared D and Dhariwal, Prafulla and Neelakantan, Arvind and Shyam, Pranav and Sastry, Girish and Askell, Amanda and others},
  journal={Advances in neural information processing systems},
  volume={33},
  pages={1877--1901},
  year={2020}
}

@article{chen2024text,
  title={Text-space graph foundation models: Comprehensive benchmarks and new insights},
  author={Chen, Zhikai and Mao, Haitao and Liu, Jingzhe and Song, Yu and Li, Bingheng and Jin, Wei and Fatemi, Bahare and Tsitsulin, Anton and Perozzi, Bryan and Liu, Hui and others},
  journal={Advances in Neural Information Processing Systems},
  volume={37},
  pages={7464--7492},
  year={2024}
}

@article{chen2024llaga,
  title={Llaga: Large language and graph assistant},
  author={Chen, Runjin and Zhao, Tong and Jaiswal, Ajay and Shah, Neil and Wang, Zhangyang},
  journal={arXiv preprint arXiv:2402.08170},
  year={2024}
}

@article{pei2020geom,
  title={Geom-gcn: Geometric graph convolutional networks},
  author={Pei, Hongbin and Wei, Bingzhe and Chang, Kevin Chen-Chuan and Lei, Yu and Yang, Bo},
  journal={arXiv preprint arXiv:2002.05287},
  year={2020}
}

@article{rozemberczki2021multi,
  title={Multi-scale attributed node embedding},
  author={Rozemberczki, Benedek and Allen, Carl and Sarkar, Rik},
  journal={Journal of Complex Networks},
  volume={9},
  number={2},
  pages={cnab014},
  year={2021},
  publisher={Oxford University Press}
}

@article{platonov2023critical,
  title={A critical look at the evaluation of GNNs under heterophily: Are we really making progress?},
  author={Platonov, Oleg and Kuznedelev, Denis and Diskin, Michael and Babenko, Artem and Prokhorenkova, Liudmila},
  journal={arXiv preprint arXiv:2302.11640},
  year={2023}
}

@article{zhu2020beyond,
  title={Beyond homophily in graph neural networks: Current limitations and effective designs},
  author={Zhu, Jiong and Yan, Yujun and Zhao, Lingxiao and Heimann, Mark and Akoglu, Leman and Koutra, Danai},
  journal={Advances in neural information processing systems},
  volume={33},
  pages={7793--7804},
  year={2020}
}

@inproceedings{muller2022transformers,
title={Transformers Can Do Bayesian Inference},
author={Samuel M{\"u}ller and Noah Hollmann and Sebastian Pineda Arango and Josif Grabocka and Frank Hutter},
booktitle={International Conference on Learning Representations},
year={2022},
url={https://openreview.net/forum?id=KSugKcbNf9}
}

@inproceedings{li2021deeprobust,
  title={Deeprobust: a platform for adversarial attacks and defenses},
  author={Li, Yaxin and Jin, Wei and Xu, Han and Tang, Jiliang},
  booktitle={AAAI},
  pages={16078--16080},
  year={2021}
}

@incollection{bongini2022biognn,
  title={BioGNN: how graph neural networks can solve biological problems},
  author={Bongini, Pietro and Pancino, Niccol{\`o} and Scarselli, Franco and Bianchini, Monica},
  booktitle={Artificial Intelligence and Machine Learning for Healthcare: Vol. 1: Image and Data Analytics},
  pages={211--231},
  year={2022},
  publisher={Springer}
}

@article{kong2024gofa,
  title={Gofa: A generative one-for-all model for joint graph language modeling},
  author={Kong, Lecheng and Feng, Jiarui and Liu, Hao and Huang, Chengsong and Huang, Jiaxin and Chen, Yixin and Zhang, Muhan},
  journal={arXiv preprint arXiv:2407.09709},
  year={2024}
}

@article{wu2025graph,
  title={Graph-R1: Incentivizing the Zero-Shot Graph Learning Capability in LLMs via Explicit Reasoning},
  author={Wu, Yicong and Lu, Guangyue and Zuo, Yuan and Zhang, Huarong and Wu, Junjie},
  journal={arXiv preprint arXiv:2508.17387},
  year={2025}
}

@inproceedings{li2024zerog,
  title={Zerog: Investigating cross-dataset zero-shot transferability in graphs},
  author={Li, Yuhan and Wang, Peisong and Li, Zhixun and Yu, Jeffrey Xu and Li, Jia},
  booktitle={Proceedings of the 30th ACM SIGKDD Conference on Knowledge Discovery and Data Mining},
  pages={1725--1735},
  year={2024}
}

@inproceedings{qiu2018deepinf,
  title={Deepinf: Social influence prediction with deep learning},
  author={Qiu, Jiezhong and Tang, Jian and Ma, Hao and Dong, Yuxiao and Wang, Kuansan and Tang, Jie},
  booktitle={Proceedings of the 24th ACM SIGKDD international conference on knowledge discovery \& data mining},
  pages={2110--2119},
  year={2018}
}

@article{erdds1959random,
  title={On random graphs I},
  author={Erd{\H{o}}s, Paul and R{\'e}nyi, Alfr{\'e}d},
  journal={Publ. math. debrecen},
  volume={6},
  number={290-297},
  pages={18},
  year={1959}
}

@article{holland1983stochastic,
  title={Stochastic blockmodels: First steps},
  author={Holland, Paul W and Laskey, Kathryn Blackmond and Leinhardt, Samuel},
  journal={Social networks},
  volume={5},
  number={2},
  pages={109--137},
  year={1983},
  publisher={Elsevier}
}

@inproceedings{zhu2002learning,
  title={Learning from labeled and unlabeled data with label propagation},
  author={Zhu, Xiaojin and Ghahramani, Zoubin},
  booktitle={Technical Report CMU-CALD-02-107, Carnegie Mellon University},
  year={2002}
}

@inproceedings{wu2019simplifying,
  title={Simplifying graph convolutional networks},
  author={Wu, Felix and Souza, Amauri and Zhang, Tianyi and Fifty, Christopher and Yu, Tao and Weinberger, Kilian Q},
  booktitle={Proceedings of the 36th International Conference on Machine Learning (ICML)},
  pages={6861--6871},
  year={2019}
}

@inproceedings{chien2021adaptive,
  title={Adaptive Universal Generalized PageRank Graph Neural Network},
  author={Chien, Eli and Peng, Jie and Li, Pan and Milenkovic, Olgica},
  booktitle={Proceedings of the 9th International Conference on Learning Representations (ICLR)},
  year={2021}
}

@inproceedings{ribeiro2017struc2vec,
  title={struc2vec: Learning node representations from structural identity},
  author={Ribeiro, Leonardo FR and Saverese, Pedro HP and Figueiredo, Daniel R},
  booktitle={Proceedings of the 23rd ACM SIGKDD international conference on knowledge discovery and data mining},
  pages={385--394},
  year={2017}
}

@inproceedings{cui2022positional,
  title={On positional and structural node features for graph neural networks on non-attributed graphs},
  author={Cui, Hejie and Lu, Zijie and Li, Pan and Yang, Carl},
  booktitle={Proceedings of the 31st ACM International Conference on Information \& Knowledge Management},
  pages={3898--3902},
  year={2022}
}

@inproceedings{platonov2023a,
title={A critical look at the evaluation of {GNN}s under heterophily: Are we really making progress?},
author={Oleg Platonov and Denis Kuznedelev and Michael Diskin and Artem Babenko and Liudmila Prokhorenkova},
booktitle={The Eleventh International Conference on Learning Representations },
year={2023},
url={https://openreview.net/forum?id=tJbbQfw-5wv}
}

@inproceedings{nagler2023statistical,
  title={Statistical foundations of prior-data fitted networks},
  author={Nagler, Thomas},
  booktitle={International Conference on Machine Learning},
  pages={25660--25676},
  year={2023},
  organization={PMLR}
}

@article{dooley2023forecastpfn,
  title={Forecastpfn: Synthetically-trained zero-shot forecasting},
  author={Dooley, Samuel and Khurana, Gurnoor Singh and Mohapatra, Chirag and Naidu, Siddartha V and White, Colin},
  journal={Advances in Neural Information Processing Systems},
  volume={36},
  pages={2403--2426},
  year={2023}
}

@article{luan2023graph,
  title={When do graph neural networks help with node classification? investigating the homophily principle on node distinguishability},
  author={Luan, Sitao and Hua, Chenqing and Xu, Minkai and Lu, Qincheng and Zhu, Jiaqi and Chang, Xiao-Wen and Fu, Jie and Leskovec, Jure and Precup, Doina},
  journal={Advances in Neural Information Processing Systems},
  volume={36},
  pages={28748--28760},
  year={2023}
}

@article{mcpherson2001birds,
  title={Birds of a feather: Homophily in social networks},
  author={McPherson, Miller and Smith-Lovin, Lynn and Cook, James M},
  journal={Annual review of sociology},
  volume={27},
  number={1},
  pages={415--444},
  year={2001},
  publisher={Annual Reviews 4139 El Camino Way, PO Box 10139, Palo Alto, CA 94303-0139, USA}
}

@misc{sato_training-free_2024,
    title = {Training-free {Graph} {Neural} {Networks} and the {Power} of {Labels} as {Features}},
    url = {http://arxiv.org/abs/2404.19288},
    doi = {10.48550/arXiv.2404.19288},
    urldate = {2025-07-21},
    publisher = {arXiv},
    author = {Sato, Ryoma},
    month = aug,
    year = {2024},
    note = {arXiv:2404.19288 [cs]},
}

@book{peters2017elements,
  title={Elements of causal inference: foundations and learning algorithms},
  author={Peters, Jonas and Janzing, Dominik and Sch{\"o}lkopf, Bernhard},
  year={2017},
  publisher={The MIT press}
}

@book{pearl2009causality,
  title={Causality},
  author={Pearl, Judea},
  year={2009},
  publisher={Cambridge university press}
}

@article{binkiewicz2017covariate,
  title={Covariate-assisted spectral clustering},
  author={Binkiewicz, Norbert and Vogelstein, Joshua T and Rohe, Karl},
  journal={Biometrika},
  volume={104},
  number={2},
  pages={361--377},
  year={2017},
  publisher={Oxford University Press}
}

@misc{deepl,
  author = {DeepL\;SE},
  title = {DeepL Translate: The world’s most accu-
rate translator},
  year = {2025},
  note = {Last accessed 25 Sep 2026},
  url = {https://www.deepl.com/translator}
}

@misc{gtranslate,
  author = {Google\;LLC},
  title = {Google Translate},
  year = {2025},
  note = {Last accessed 25 Sep 2026},
  url = {https://translate.google.com/}
}

@inproceedings{zhu2020h2gcn,
  title={Beyond Homophily in Graph Neural Networks: Current Limitations and Effective Designs},
  author={Zhu, Jiong and Yan, Yujun and Zhao, Lingxiao and Heimann, Mark and Akoglu, Leman and Koutra, Danai},
  booktitle={Advances in Neural Information Processing Systems},
  year={2020}
}

@article{li2024graph,
  title={Graph learning in the era of llms: A survey from the perspective of data, models, and tasks},
  author={Li, Xunkai and Wu, Zhengyu and Wu, Jiayi and Cui, Hanwen and Jia, Jishuo and Li, Rong-Hua and Wang, Guoren},
  journal={arXiv preprint arXiv:2412.12456},
  year={2024}
}

@article{luan2022revisiting,
  title={Revisiting heterophily for graph neural networks},
  author={Luan, Sitao and Hua, Chenqing and Lu, Qincheng and Zhu, Jiaqi and Zhao, Mingde and Zhang, Shuyuan and Chang, Xiao-Wen and Precup, Doina},
  journal={Advances in neural information processing systems},
  volume={35},
  pages={1362--1375},
  year={2022}
}

@article{zhao2023graphtext,
  title={Graphtext: Graph reasoning in text space},
  author={Zhao, Jianan and Zhuo, Le and Shen, Yikang and Qu, Meng and Liu, Kai and Bronstein, Michael and Zhu, Zhaocheng and Tang, Jian},
  journal={arXiv preprint arXiv:2310.01089},
  year={2023}
}

@article{ye2023instructglm,
  title={Language is all a graph needs},
  author={Ye, Ruosong and Zhang, Caiqi and Wang, Runhui and Xu, Shuyuan and Zhang, Yongfeng},
  journal={arXiv preprint arXiv:2308.07134},
  year={2023}
}

@article{barabasi1999emergence,
  title={Emergence of scaling in random networks},
  author={Barab{\'a}si, Albert-L{\'a}szl{\'o} and Albert, R{\'e}ka},
  journal={science},
  volume={286},
  number={5439},
  pages={509--512},
  year={1999},
  publisher={American Association for the Advancement of Science}
}

@article{choi2025tabpfngn,
  title={Can TabPFN Compete with GNNs for Node Classification via Graph Tabularization?},
  author={Choi, Jeongwhan and Kang, Woosung and Kim, Minseo and Kim, Jongwoo and Park, Noseong},
  journal={arXiv preprint arXiv:2512.08798},
  year={2025}
}

@inproceedings{kim2025leveraging,
  title={Leveraging multi-facet paths for heterogeneous graph representation learning},
  author={Kim, Jongwoo and Chu, Seongyeub and Park, Hyeongmin and Wong, Bryan and Han, Keejun and Yi, Mun Yong},
  booktitle={Proceedings of the 34th ACM International Conference on Information and Knowledge Management},
  pages={1334--1343},
  year={2025}
}

@article{choi2025fn,
  title={Are Graph Transformers Necessary? Efficient Long-Range Message Passing with Fractal Nodes in MPNNs},
  author={Choi, Jeongwhan and Park, Seungjun and Park, Sumin and Cho, Sung-Bae and Park, Noseong},
  journal={arXiv preprint arXiv:2511.13010},
  year={2025}
}
\bibliographystyle{iclr2026_conference}
\clearpage
\appendix

\part{\Large{Supplementary Materials for ``Learning Posterior Predictive Distributions for Node Classification from Synthetic Graph Priors''}} 

\section{Details of Datasets}\label{app:dataset}

\paragraph{The Synthetic Cora Network}
The synthetic Cora dataset is provided by ~\citep{zhu2020h2gcn}. \citet{zhu2020h2gcn} generate graphs for a target homophily level using a modified preferential attachment process. We sample nodes, edges, and features from Cora to create a synthetic graph with a desired homophily and feature/label distribution. In Table~\ref{tab:syn_cora}, we summarize the properties of the synthetic Cora networks we used. 

\begin{table}[h!]
    \small
    \centering
    \caption{The detailed information of the synthetic Cora. All levels of homophily have the same number of features (1,433), nodes (1,480), edges (5,936), and classes (5).}
    \begin{tabular}{c ccc}\toprule
        Homophily & Avg. Degree & Max. Degree & Min. Degree\\ \midrule
        0.0 & 3.98 & 84.33 & 1.67 \\
        0.1 & 3.98 & 71.33 & 2.00 \\
        0.2 & 3.98 & 73.33 & 1.67 \\
        0.3 & 3.98 & 70.00 & 2.00 \\
        0.4 & 3.98 & 77.67 & 2.00 \\
        0.5 & 3.98 & 76.33 & 2.00 \\
        0.6 & 3.98 & 76.00 & 1.67 \\
        0.7 & 3.98 & 67.67 & 2.00 \\
        0.8 & 3.98 & 58.00 & 1.67 \\
        0.9 & 3.98 & 58.00 & 1.67 \\
        1.0 & 3.98 & 51.00 & 2.00 \\
        \bottomrule
    \end{tabular}
    \label{tab:syn_cora}
\end{table}

\paragraph{Real-world Graph Datasets.}
We list the dataset statistics we used in~\Cref{tab:data_homo,tab:data_hete}. 
We use 23 benchmark datasets for node classification. Following prior work, these include both 13 homophily graphs~\citep{kipf2017GCN,rozemberczki2021multi} (e.g., Cora, Citeseer, Pubmed, WikiCS) and 10 heterophilous graphs~\citep{pei2020geom,platonov2023critical} (e.g., Cornell, Texas, Squirrel, Actor). For Chameleon and Squirrel, we use filtered datasets from~\citet{platonov2023a}.
We also report the clustering coefficients of each graph dataset.

\begin{table}[h]
    \centering
    \small
    \setlength{\tabcolsep}{2pt}
    \caption{Homophily dataset statistics for node classification 13 benchmarks.}
    \label{tab:data_homo}
    \begin{tabular}{lccccccc}
    \toprule
    Dataset & \#Nodes & \#Edges & \#Features & \#Classes & \#Labels & Coeff. & Train/Val/Test (\%) \\
    \midrule
    AirBrazil    & 131   & 1,074   & N/A  & 4  & 80  & 0.6364 & 61.1/19.1/19.8 \\
    AirEU        & 1,190 & 5,995   & N/A  & 4  & 80  & 0.5393 & 20.1/39.8/40.1 \\
    AirUS        & 10,008 & 13,599 & N/A  & 4  & 80  & 0.5011 & 6.7/46.6/46.6 \\
    Cora         & 2,708  & 10,556 & 1,433 & 7  & 140 & 0.2407 & 5.2/18.5/36.9 \\
    Citeseer     & 3,327  & 9,104  & 3,703 & 6  & 120 & 0.1435 & 3.6/15.0/30.1 \\
    Pubmed       & 19,717 & 88,648 & 500   & 3  & 60  & 0.0602 & 0.3/2.5/5.1 \\
    WikiCS       & 11,701 & 431,206 & 300  & 10 & 580 & 0.4660 & 5.0/15.1/49.9 \\
    Amazon-Photo & 7,650  & 238,162 & 745  & 8  & 160 & 0.4101 & 2.1/49.0/49.0 \\
    Amazon-Comp  & 13,752 & 491,722 & 767  & 10 & 200 & 0.3513 & 1.5/49.3/49.3 \\
    DBLP         & 17,716 & 105,734 & 1,639 & 4  & 80  & 0.1344 & 0.5/49.8/49.8 \\
    Coauthor-CS  & 18,333 & 163,788 & 6,805 & 15 & 300 & 0.3425 & 1.6/49.2/49.2 \\
    Coauthor-Physics & 34,493 & 495,924 & 8,415 & 5  & 100 & 0.3776 & 0.3/49.9/49.9 \\
    Deezer       & 28,281 & 185,504 & 128   & 2  & 40  & 0.1412 & 0.1/49.9/49.9 \\
    \bottomrule
    \end{tabular}
\end{table}

\begin{table}[h!]
    \centering
    \small
    \setlength{\tabcolsep}{2pt}
    \caption{Heterophily dataset statistics for 10 benchmarks.}
    \label{tab:data_hete}
    \begin{tabular}{lccccccc}
    \toprule
    Dataset & \#Nodes & \#Edges & \#Features & \#Classes & \#Labels & Coeff. & Train/Val/Test (\%) \\
    \midrule
    Cornell       & 183   & 554     & 1703 & 5  & 87    & 0.1671 & 47.5/32.2/20.2 \\
    Texas         & 183   & 558     & 1703 & 5  & 87    & 0.1979 & 47.5/31.7/20.2 \\
    Wisconsin     & 251   & 900     & 1703 & 5  & 120   & 0.2077 & 47.8/31.9/20.3 \\
    Chameleon     & 890  & 8854  & 2325 & 5  & 445  & 0.5769 & 50.0/25.0/25.0 \\
    Actor         & 7600  & 30,019  & 932  & 5  & 3648  & 0.0802 & 48.0/32.0/20.0 \\
    Minesweeper   & 10,000& 78,804  & 7    & 2  & 5000  & 0.4355 & 50.0/25.0/25.0 \\
    Tolokers      & 11,758& 1,038,000 & 10 & 2  & 5879  & 0.5329 & 50.0/25.0/25.0 \\
    Amazon-Ratings  & 24,492& 186,100 & 300  & 5  & 12,246 & 0.5816 & 50.0/25.0/25.0 \\
    Questions     & 48,921& 307,080 & 301  & 2  & 24,460 & 0.0307 & 50.0/25.0/25.0 \\
    Squirrel      & 2223  & 46,998 & 2089 & 5  & 1053  & 0.4631 & 50.0/25.0/25.0 \\
    \bottomrule
    \end{tabular}
\end{table}

\section{Detailed Experimental Settings}\label{app:setting}
\subsection{Hardware and Software Specifications}
Our implementation is developed on top of the \textsc{TabPFN-v1}\footnote{\url{github.com/PriorLabs/TabPFN/tree/tabpfn_v1/}} framework.
All experiments were performed using the following software and hardware environments: \textsc{Ubuntu} 21.04 LTS, \textsc{Python} 3.10.16, \textsc{PyTorch} 1.12.1, \textsc{PyTorch Geometric} 2.3.1, \textsc{torch-scatter} 2.1.2, \textsc{torch-sparse} 0.6.18, \textsc{Numpy} 1.26.4, \textsc{networkx} 3.3, \textsc{scikit-learn} 1.4.0, \textsc{CUDA} 12.3, \textsc{NVIDIA} Driver 550.54.14, i9 CPU, \textsc{NVIDIA} RTX 6000.

\subsection{Training Setup}
The model configuration of NodePFN is summarized in Table~\ref{tab:model-config}.  In total, the model contains approximately 29.1M trainable parameters.
We trained NodePFN for a total of 30 epochs, each epoch comprising up to 1,024 steps (245,760 steps in total) with a batch size of 8 (see more hyperparameters in \Cref{tab:train-hparams}. 
The total training required approximately 6 GPU hours on a single \textsc{NVIDIA} RTX A6000.

\begin{table}[h!]
    \centering
    \caption{Model configuration of NodePFN.}
    \label{tab:model-config}
    \small
    \begin{tabular}{ll}
    \toprule
    Hyperparameter & Value \\
    \midrule
    Embedding dimension  & 512 \\
    Number of layers           & 12 \\
    Number of attention heads  & 4 \\
    Dropout rate               & 0.0 \\
    \bottomrule
    \end{tabular}
\end{table}
\begin{table}[h]
    \centering
    \caption{Training hyperparameters for NodePFN.}
    \label{tab:train-hparams}
    \small
    \begin{tabular}{ll}
    \toprule
    Hyperparameter & Value \\
    \midrule
    Epochs                  & 30 \\
    Steps per epoch         & 1024 \\
    Batch size              & 8 \\
    Embedding size          & 512 \\
    Learning rate           & $\{1.5 \times 10^{-5},\, 5 \times 10^{-4},\, 1 \times 10^{-4}\}$ \\
    Optimizer               & AdamW \\
    \bottomrule
    \end{tabular}
\end{table}

\subsection{Details of NodePFN Prior}\label{app:prior}
\paragraph{Structural Causal Models (SCM)}
We adopt the optimal sampling distributions from TabPFN~\citep{hollmann2023tabpfn} for our SCM prior to ensure robust feature-label relationships. 
Following TabPFN's framework\footnote{\url{https://github.com/PriorLabs/TabPFN/tree/tabpfn_v1}}, each SCM is generated by:
\begin{itemize}
    \item Sampling MLP layers $\ell_{\mathrm{scm}}\sim p(\ell_{\mathrm{scm}})$ and hidden size $h_{\mathrm{scm}}\sim p(h_{\mathrm{scm}})$ from discretized noisy log-normal distributions
    \item Creating a layered graph structure and randomly dropping edges to form a DAG
    \item Selecting feature nodes and one label node from the causal graph
    \item Using activation functions sampled from {Tanh, LeakyReLU, ELU, Identity}
    \item Applying noisy log-normal noise distributions with beta-distributed dropout rates
\end{itemize}
This generates complex non-linear feature-label dependencies while maintaining the causal structure that has proven effective for tabular data modeling.
\paragraph{Contextual SBM.}
Our contextual SBM generates community-structured graphs with controllable homophily levels:
\begin{itemize}
    \item Sample homophily rate $h \sim \mathcal{U}(0.1, 0.9)$ and intra-community probability $p_{\mathrm{in}} \sim \mathcal{U}(0.01, 0.1)$
    \item Compute inter-community probability as $p_{\mathrm{out}} = p_{\mathrm{in}} \times (1 - h)$
    \item Assign nodes to communities based on their labels from the SCM
    \item Generate edge probabilities using power distributions: $\mathrm{probs}_{i,j} = \mathrm{Power}(5) \times p_{\mathrm{out}} \quad \mathrm{for } i \neq j$ for inter-community edges, with diagonal values $p_{\mathrm{out}} + \mathrm{power}(2, size) × (p_{\mathrm{in}} - p_{\mathrm{out}})$ for intra-community connections
    \item Create a symmetric probability matrix and generate edges via the stochastic block model
\end{itemize}
This approach ensures that network topology and node labels are inherently related through the homophily parameter.
\paragraph{ER Network.}
ER networks provide unstructured baseline graphs complementing the community-based patterns:
\begin{itemize}
    \item Sample edge probability $p_{\mathrm{er}} \sim \mathcal{U}(0.01, 0.05)$
    \item Generate edges independently with probability $\mathcal{E}_{ij} \sim \mathrm{Bernoulli}(p_{\mathrm{er}})$
    \item Creates graphs with binomial degree distributions and no inherent community structure
\end{itemize}
The combination of cSBMs (50\% of training graphs) and ER networks (50\% of training graphs) ensures comprehensive coverage of graph structures from community-based patterns to random connectivity. 

\subsection{Flexible Encoder for Variable Node Feature Dimensions}\label{app:flex_encoder}
Our NodePFN model is designed to handle graphs with varying node feature dimensionalities up to a pre-defined maximum capacity. This is achieved through a flexible input encoder that standardizes the feature vectors before they are processed by the main architecture. 

When a graph is provided where the node features have a dimension $d$ that is less than the maximum supported dimension of the model, $d_{max}$, each node's feature vector is first extended to $d_{max}$ dimensions by appending zero-padding. 
Then, to ensure that the zero-padding process does not systematically alter the input's scale or introduce bias, we apply a normalization factor to the padded vector. This mechanism ensures that feature vectors from different graph datasets are processed on a consistent scale, enabling our single pre-trained model to generalize effectively across a wide range of graph-structured data.

\subsection{Feature and Label Embeddings in Implementation}\label{app:imp-emb}
We employ learnable linear transformations $\mathbf{W}_{\mathcal{X}} \in \mathbb{R}^{d \times d_{\text{feat}}}$ and $\mathbf{W}_{\mathcal{Y}} \in \mathbb{R}^{d \times C}$ to project features and labels to the embedding dimension $d$, where $d_{\text{feat}}$ is the feature dimension and $C$ is the number of classes. Following TabPFN-v1's implementation, we use element-wise addition rather than concatenation to combine features and labels:
\begin{align}
\mathbf{H}_{\text{train}}^{(0)} = \mathbf{X}_{\text{train}}\mathbf{W}_{\mathcal{X}}^{\top} + \mathbf{Y}_{\text{train}}\mathbf{W}_{\mathcal{Y}}^{\top} ,\;\;
\mathbf{H}_{\text{test}}^{(0)} = \mathbf{X}_{\text{test}}\mathbf{W}_{\mathcal{X}}^{\top},\nonumber
\end{align}
where $\mathbf{X}_{\text{train}}$ and $\mathbf{Y}_{\text{train}}$ are the feature and label matrices for training nodes. This additive approach\footnote{\url{https://github.com/PriorLabs/TabPFN/issues/93}} maintains constant dimensionality and enables the model to learn complementary representations in different subspaces of the embedding vector. During inference, real-world labels are first converted to canonical integers before applying $\mathbf{W}_{\mathcal{Y}}$.

\subsection{Inference Implementation Details}\label{app:inference}
We describe the data preprocessing methods used in the inference stage below. 
\paragraph{Graph Structure Preprocessing.}
We convert the adjacency matrix of a given graph into a normalized adjacency matrix.
\paragraph{Feature Preprocessing.}
Adopting the ensemble approach from \citet{hollmann2023tabpfn}, we employ a method that alters the order and scaling of features within the input context. This integrates a form of ensemble technique, using a fixed number of 32 multiple inputs, with the subsequent prediction results being aggregated.
When real-world graphs have features exceeding the capacity of our NodePFN, we apply truncated SVD for dimensionality reduction. We also optionally apply feature smoothing using sum aggregation by edge connectivity for enhanced feature quality.

\subsection{Experimental Settings for Node Classification}\label{app:hyp_nc}

\paragraph{Evaluation Protocol.}
For homophily datasets, we follow the semi-supervised setting of \citet{kipf2017GCN} for Cora, Citeseer, and Pubmed (20 nodes per class for training, 500 validation, 1000 test), while for WikiCS,  we follow the splits in \citet{rozemberczki2021multi}, and the remaining datasets follow the GraphAny~\citep{zhao2025graphany} protocol (20 nodes per class for training, the rest split evenly into validation and test).
For heterophily datasets, we use the predefined split masks provided in \citet{pei2020geom} and \citet{platonov2023critical}.
For Chameleon and Squirrel, we use filtered datasets from~\citet{platonov2023a}.

\paragraph{Search Space of Hyperparameters.}
We report our search space of hyperparameters used in our experiments in \Cref{tab:hyp_range}.
Note that we use the default hyperparameter of 32 ensembles.
\begin{table}[h!]
    \small
    \centering
    \caption{Homophily dataset hyperparameters for node classification.}
    \label{tab:hyp_range}
    \begin{tabular}{lc}
    \toprule
    Hyperparam. & Range \\
    \midrule
    \# Components      & \{10,15,20,25,30\}\\
    \# Smoothing Steps & \{0, 1, 2, 3, 4\}\\
    \bottomrule
    \end{tabular}
\end{table}

\paragraph{Optimal Hyperparameters.}
We report our optimal hyperparameters used in our experiments in \Cref{tab:homo_hyp,tab:hete_hyp}. We do not use truncated SVD on Tolokers and Minesweeper.

\begin{table}[h!]
    \small
    \centering
    \caption{Homophily dataset hyperparameters for node classification.}
    \label{tab:homo_hyp}
    \begin{tabular}{lcc}
    \toprule
    Dataset & \# Components & \# Smoothing Steps\\
    \midrule
    AirBrazil         & 25 & 3 \\
    AirEU             & 25 & 1 \\
    AirUS             & 25 & 3 \\
    Cora              & 15 & 4 \\
    Citeseer          & 15 & 2 \\
    Pubmed            & 15 & 2 \\
    WikiCS            & 15 & 2 \\
    Amazon-Photo      & 15 & 3 \\
    Amazon-Comp       & 15 & 3 \\
    DBLP              & 25 & 2 \\
    Coauthor-CS       & 25 & 2 \\
    Coauthor-Physics  & 15 & 0 \\
    Deezer            & 25 & 2 \\
    \bottomrule
    \end{tabular}
\end{table}

\begin{table}[h]
    \centering
    \small
    \caption{Heterophily dataset hyperparameters for node classification.}
    \label{tab:hete_hyp}
    \begin{tabular}{lcc}
    \toprule
    Dataset & \# Components & \# Smoothing steps\\
    \midrule
    Cornell     & 15  & 0 \\
    Texas       & 20  & 0 \\
    Wisconsin   & 25  & 0 \\
    Chameleon   & 25  & 0 \\
    Actor       & 35 & 0 \\
    Minesweeper & -  & 1 \\
    Tolokers    & -  & 2 \\
    Amazon-Ratings & 25  & 3 \\
    Questions   & 25  & 3   \\
    Squirrel    & 15  & 1  \\
    \bottomrule
    \end{tabular}
\end{table}

\clearpage
\section{Prior Data Scale Analysis}\label{app:prior_scale}

\begin{wrapfigure}{r}{0.4\textwidth}
    \vspace{-1em}
    \includegraphics[width=0.4\textwidth]{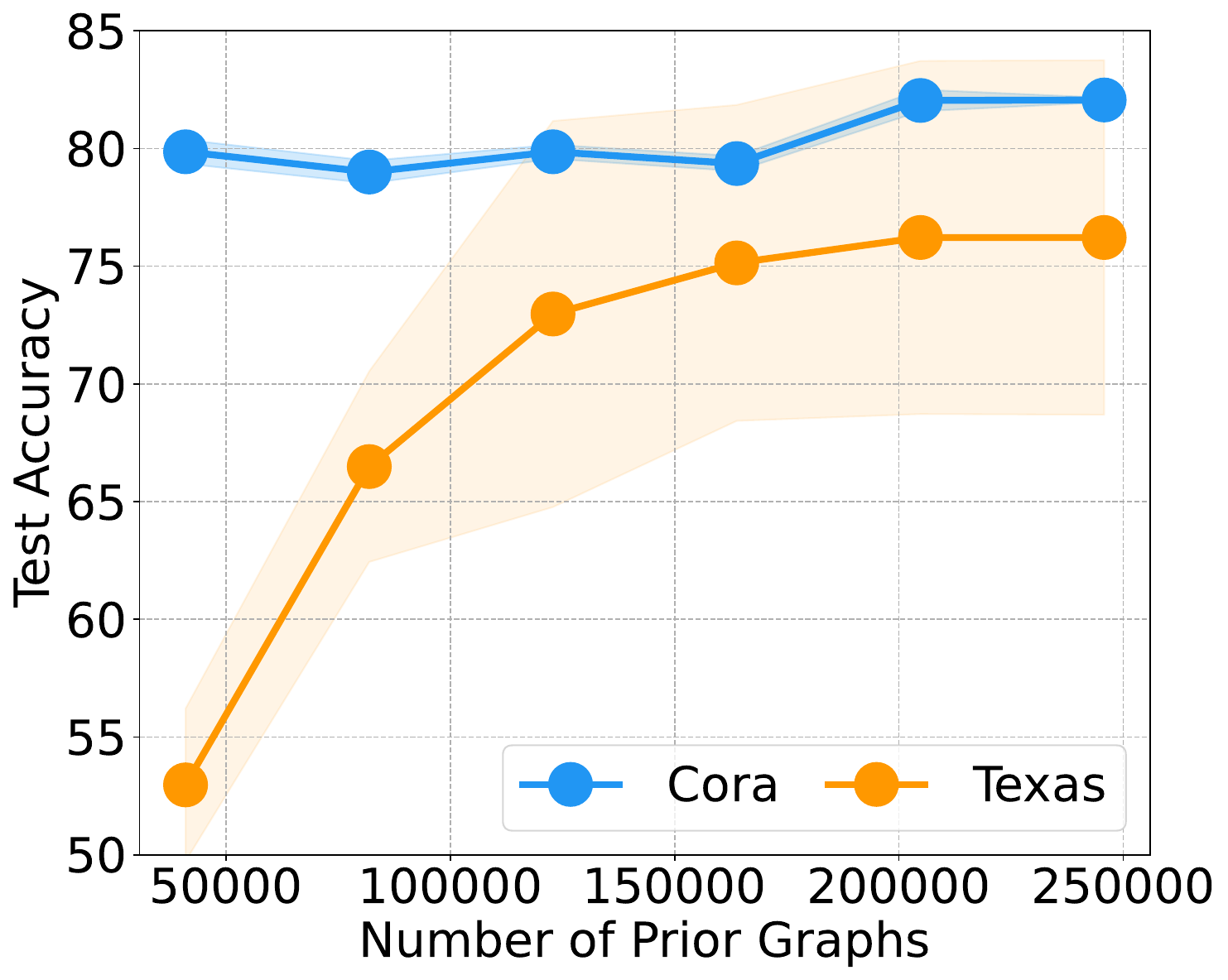}
    \caption{Impact of Prior Data Scale on NodePFN Performance}
    \label{fig:scale}
    \vspace{-1em}
\end{wrapfigure}
We analyze the impact of the number of synthetic prior graphs on NodePFN performance and address the question of data efficiency in PFN training. This analysis aims to understand the trade-off between computational cost and performance gains when scaling prior data.

In our pretraining, each training iteration generates new synthetic graphs based on our random network prior, which creates diverse structural patterns. \Cref{fig:scale} reveals several key insights: (1) Particularly evident in the heterophily Texas dataset, its performance increases substantially with more prior data. The accuracy improves from approximately 53\% to 76\%. (2) For Cora, NodePFN shows more modest gains, and this suggests that patterns with higher homophily rates are effectively learned via cSBMs. (3) The performance improvement peaks at approximately 250,000 prior graphs. 

While generating synthetic prior data incurs an initial computational cost, this expense is amortized across all future inference tasks. Unlike traditional GNNs, which require retraining for each new graph, NodePFN enables immediate inference on arbitrary real-world graphs via one-time training on synthetic priors.  Given the universal applicability of the resulting model, the computational investment in comprehensive prior generation proves worth it.

\section{Comparison with TabPFN and NodePFN}\label{app:vs_tabpfn}
In \Cref{tab:tabpfn_vs_nodepfn_homo,tab:tabpfn_vs_nodepfn_heto}, we report all results from \Cref{fig:violin} in \Cref{sec:ablation}.
We used the TabPFN-v1 framework for comparison with TabPFN, and to ensure a fair comparison, we employed a pre-trained model using the same number of prior data points as ours. 
For TabPFN, we use the default hyperparameter of 32 ensembles, and we also enable feature subsampling for feature preprocessing.

\begin{table}[htbp]
\begin{minipage}{.49\textwidth}
\centering
\footnotesize
\setlength{\tabcolsep}{5pt}
\renewcommand{\arraystretch}{1.2}
\caption{Comparison between TabPFN and NodePFN on homophily datasets (accuracy \%).}
\label{tab:tabpfn_vs_nodepfn_homo}
\begin{tabular}{lcc}
\toprule
Dataset & TabPFN & NodePFN \\
\midrule
AirBrazil        & 52.31 & 75.38 \\
AirEU            & 53.62 & 57.00 \\
AirUS            & 53.55 & 61.66 \\
Cora             & 53.10 & 82.06 \\
Citeseer         & 32.94 & 67.30 \\
Pubmed           & 65.04 & 78.00 \\
WikiCS           & 62.25 & 75.98 \\
Amazon-Photo     & 74.77 & 90.53 \\
Amazon-Comp      & 50.66 & 81.42 \\
DBLP             & 60.69 & 74.71 \\
Co-author CS     & 48.32 & 91.55 \\
Co-author Physics& 63.41 & 93.43 \\
Deezer           & 49.83 & 53.45 \\
\midrule
\textbf{Average} & 55.62 & 77.39 \\
\bottomrule
\end{tabular}
\end{minipage}
\begin{minipage}{.49\textwidth}
\centering
\footnotesize
\setlength{\tabcolsep}{5pt}
\renewcommand{\arraystretch}{1.2}
\label{tab:tabpfn_vs_nodepfn_homo}
\centering
\small
\setlength{\tabcolsep}{5pt}
\renewcommand{\arraystretch}{1.2}
\caption{Comparison between TabPFN and NodePFN on heterophily datasets (accuracy \%).}
\label{tab:tabpfn_vs_nodepfn_heto}
\begin{tabular}{lcc}
\toprule
Dataset & TabPFN & NodePFN \\
\midrule
Cornell        & 55.68 & 71.89 \\
Texas          & 62.70 & 76.22 \\
Wisconsin      & 72.94 & 79.22 \\
Chameleon      & 37.54 & 50.13 \\
Actor          & 25.84 & 32.99 \\
Minesweeper    & 79.86 & 80.66 \\
Tolokers       & 78.18 & 78.61 \\
Amazon-Ratings & 21.64 & 44.68 \\
Questions      & 90.09 & 97.02 \\
Squirrel       & 31.84 & 43.40 \\
\midrule
\textbf{Average} & 53.47 & 65.14 \\
\bottomrule
\end{tabular}
\end{minipage}
\end{table}

\clearpage

\section{Theoretical Discussion}\label{app:theo}
The original PFN framework \citep{muller2022transformers} establishes that Transformers can approximate posterior predictive distributions (PPDs) by minimizing the Prior-Data negative log-likelihood. 
Specifically, \citet[Insight 1]{muller2022transformers} shows that this loss equals the expected cross-entropy between the true PPD and its approximation, while \citet[Corollary 1.2]{muller2022transformers} guarantees convergence to the exact posterior under the infinite-capacity assumption, provided the architecture respects the exchangeability of the conditioning dataset $\mathcal{D}$. In practice, this requires the architecture to be permutation equivariant with respect to the ordering of training examples.

\begin{remark}[Preservation of PFN Guarantees in NodePFN]
The dual-branch architecture of NodePFN maintains permutation equivariance because: the attention branch uses self-attention and cross-attention operations that are inherently permutation equivariant, the MPNN branch uses aggregation functions that are permutation equivariant, and their additive fusion (\Cref{eq:fusion}) preserves this property. Therefore, NodePFN satisfies the exchangeability requirement of \citet[Corollary 1.2]{muller2022transformers} and converges to the posterior.
\end{remark}
This theoretical guarantee ensures that while the MPNN branch enriches the model with structural information, the fundamental Bayesian convergence properties of PFN remain intact.

\section{Comparison with Heterophily-Specific GNNs}\label{app:vs_het}
To further validate NodePFN's effectiveness on heterophily graphs, we compare against GNNs specifically designed for heterophily: H2GCN~\citep{zhu2020h2gcn}, GPRGNN~\citep{chien2021GPRGNN}, and FAGCN~\citep{Bo2021fagcn}.
As shown in Table \ref{tab:nodepfn_vs_gnn_hete}, NodePFN achieves the best performance on 7 out of 9 datasets despite using no real-world training data, while these specialized methods require dataset-specific training with carefully designed aggregation schemes for heterophily. 

Notably, NodePFN shows improvements on Chameleon and Squirrel. The competitive performance on Texas and Actor within 1\% of the best methods further confirms that learning from diverse synthetic graphs with controlled homophily provides generalization on the heterophily spectrum without requiring architectural modifications or dataset-specific tuning.

\begin{table}[h!]
    \footnotesize
    \centering
    \setlength{\tabcolsep}{1.2pt}
    \caption{Comparison of NodePFN with H2GCN, GPRGNN, FAGCN (accuracy \%).}
    \label{tab:nodepfn_vs_gnn_hete}
    \begin{tabular}{l cc cccc ccc}\toprule
        Dataset & Chameleon & Squirrel & Cornell & Texas & Actor & Wisconsin & A.Ratings & Co.CS  & Co.Physics \\\midrule
        H2GCN  & 41.07\std{2.65} &  35.10\std{1.15} &  65.77\std{6.80} & \textbf{76.58\std{1.56}} & \textbf{35.86\std{1.03}} & 75.82\std{1.13} & 40.87\std{0.11} & 88.45\std{0.97} & 92.86\std{0.36}\\ 
        GPRGNN  & 39.69\std{1.15} &  38.95\std{1.99} & 40.54\std{2.01} & 65.77\std{1.56} & 33.94\std{0.95}  &  75.21\std{4.08} & 42.23\std{0.25} & 91.49\std{0.39} &  92.76\std{0.20}\\ 
        FAGCN & 37.24\std{3.54} & 36.78\std{3.11} & 60.38\std{1.82}  & 68.44\std{1.78} & 34.87\std{1.25} & 72.02\std{5.24} & 44.12\std{0.31} & 91.07\std{1.28} & 92.34\std{0.40}\\
        \midrule
        \textbf{NodePFN} & \textbf{50.13\std{3.30}} & \textbf{43.40\std{1.03}} & \textbf{71.89\std{2.76}} & 76.22\std{7.53} & 32.99\std{1.09} & \textbf{79.22\std{6.97}} & \textbf{44.68\std{0.48}} & \textbf{91.55\std{0.32}} & \textbf{93.43\std{0.13}}\\ 
    \bottomrule
    \end{tabular}
\end{table}

\section{Comparison with LLM-based Graph Methods on GLBench}
Following the experimental setting of recent LLM-based graph methods, we conduct the supervised node classification experiments on all the datasets in GLBench~\citep{li2024glbench}~\footnote{\url{https://github.com/NineAbyss/GLBench}}.

NodePFN achieves competitive or superior performance compared to LLM-based graph foundation models without requiring text descriptions or language model dependencies. While LLM-based methods leverage pre-trained language knowledge, NodePFN leverages pre-trained patterns from massive synthetic prior data.

\begin{table}[h!]
    \footnotesize
    \centering
    \caption{Accuracy under the supervised setting of GLBench~\citep{li2024glbench}. \textbf{Best} and \underline{second-best} are highlighted.}
    \label{tab:glbench}
    \begin{tabular}{l cccc }\toprule
        Dataset & Cora & Citeseer & Pubmed & WikiCS \\\midrule
        InstructGLM~\citep{ye2023instructglm}     & 69.10 & 51.87 & 71.26 & 45.73 \\
        GraphText~\citep{zhao2023graphtext}       & \underline{76.21} & 59.43 & \underline{75.11} & 67.35 \\
        LLaGA~\citep{chen2024llaga}& 74.42 & 55.73 & 68.82 & 73.88 \\
        OFA~\citep{liu2024ofa} & 75.24 & \textbf{73.04} & \textbf{75.61} & \underline{77.34} \\
        \cmidrule(lr){1-5}
        \textbf{NodePFN} & \textbf{76.38} & \underline{63.08 } & 68.18 & \textbf{76.29} \\
    \bottomrule
    \end{tabular}
\end{table}

\section{Comparison with TabPFN Using Smoothed Features}
\Cref{tab:tabpfn_smoothed_homo,tab:tabpfn_smoothed_hete} present comprehensive comparisons between TabPFN-v1~\citep{hollmann2023tabpfn} with smoothed features and NodePFN across homophily and heterophily datasets. 
The smoothed features baseline applies non-parametric feature aggregation before feeding node representations into TabPFN-v1's official checkpoint\footnote{\url{https://github.com/PriorLabs/TabPFN/tree/tabpfn_v1}}, effectively incorporating local neighborhood information without explicit graph structure modeling. 
On homophily datasets (\Cref{tab:tabpfn_smoothed_homo}), NodePFN demonstrates consistent improvements. The advantages become even more pronounced on heterophily datasets (\Cref{tab:tabpfn_smoothed_hete}), where NodePFN substantially outperforms the smoothed feature baseline on Cornell, Texas, and Wisconsin.
These results validate that NodePFN's explicit modeling of graph topology through its dual-branch architecture provides meaningful improvements over simple feature smoothing approaches. Note that TabPFN-v1's limitation to 10 classes prevents evaluation on datasets like Co-author CS, whereas NodePFN supports up to 20 classes.

\begin{table}[htbp]
\begin{minipage}{.49\textwidth}
\centering
\footnotesize
\setlength{\tabcolsep}{2pt}
\renewcommand{\arraystretch}{1.2}
\caption{Comparison between TabPFN with smoothed features and NodePFN on homophily datasets (accuracy \%).}
\label{tab:tabpfn_smoothed_homo}
\begin{tabular}{lcc}
\toprule
Dataset & \makecell{TabPFN-v1\\(smoothed features)}  & NodePFN \\
\midrule
AirBrazil         & 67.69 & 75.38 \\
AirEU             & 55.62 & 57.00 \\
AirUS             & 59.60 & 61.66 \\
Cora              & 74.06 & 82.06 \\
Citeseer          & 51.16 & 67.30 \\
Pubmed            & 75.96 & 78.00 \\
WikiCS            & 74.90 & 75.98 \\
Amazon-Photo      & 83.69 & 90.53 \\
Amazon-Comp       & 75.61 & 81.42 \\
DBLP              & 69.20 & 74.71 \\
Co-author CS      & N/A & 91.55 \\
Co-author Physics & 87.93 & 93.43 \\
Deezer            & 48.17 & 53.45 \\
\bottomrule
\end{tabular}
\end{minipage}
\begin{minipage}{.49\textwidth}
\centering
\footnotesize
\setlength{\tabcolsep}{2pt}
\renewcommand{\arraystretch}{1.2}
\centering
\small
\setlength{\tabcolsep}{5pt}
\renewcommand{\arraystretch}{1.2}
\caption{Comparison between TabPFN-v1 with smoothed features and NodePFN on heterophily datasets (accuracy \%).}
\label{tab:tabpfn_smoothed_hete}
\begin{tabular}{lcc}
\toprule
Dataset & \makecell{TabPFN-v1\\(smoothed features)} & NodePFN \\
\midrule
Cornell        & 42.16 & 71.89 \\
Texas          & 56.22 & 76.22 \\
Wisconsin      & 51.37 & 79.22 \\
Chameleon      & 41.42 & 50.13 \\
Actor          & 25.29 & 32.99 \\
Minesweeper    & 80.07 & 80.66 \\
Tolokers       & 78.05 & 78.61 \\
Amazon-Ratings & 44.24 & 44.68 \\
Questions      & 97.02 & 97.02 \\
Squirrel       & 40.42 & 43.40 \\
\bottomrule
\end{tabular}
\end{minipage}
\end{table}

\section{Extensive Ablations on Synthetic Prior Design}
\paragraph{Graph generation models and homophily distribution.}
Table~\ref{tab:ablation_prior_design} shows ablation results on different graph generation models and homophily distributions. We compare our approach against various alternatives, including using only ER graphs, only cSBM graphs with restricted or full homophily ranges, and only Barabási-Albert (BA) networks~\citep{barabasi1999emergence}. 
As shown in \Cref{tab:ablation_prior_design}, the results reveal several key insights. First, restricting training to specific homophily ranges leads to performance degradation. This shows the necessity of covering the full homophily spectrum to generalize across diverse real-world graphs. Second, training exclusively on Barabási-Albert networks --- which explicitly model power-law degree distributions --- shows inconsistent performance. This suggests that power-law topology alone provides insufficient structural diversity for universal graph learning. Third, using only ER or only cSBM graphs underperforms the combined approach, validating that both graph types contribute complementary inductive biases.

\begin{table}[ht!]
\centering
\small
\caption{Ablation study on graph generation models and homophily distributions.}
\label{tab:ablation_prior_design}
\begin{tabular}{lccc}
\toprule
Ablation & Cora & Wisconsin & Tolokers \\
\midrule
Only ER & 80.62 & 80.39 & 77.18 \\
Only cSBM (0.1-0.3) & 79.89 & 79.98 & 77.65 \\
Only cSBM (0.7-0.9) & 80.42 & 78.20 & 77.23 \\
Only cSBM (full range) & 81.26 & 78.82 & 77.30 \\
Only BA (Barabási-Albert) & 74.18 & 80.57 & 74.63 \\
\midrule
\textbf{NodePFN (ER+cSBM)} & \textbf{82.06} & \textbf{81.18} & \textbf{78.61} \\
\bottomrule
\end{tabular}
\end{table}

\paragraph{ER/cSBM ratio analysis.}
\Cref{tab:ablation_ratio} examines how the mixture ratio between ER and cSBM graphs affects performance. We varied the proportion of ER graphs from 0\% (cSBM only) to 100\% (ER only). The balanced 50/50 ratio consistently achieves optimal or near-optimal performance across all homophily regimes. Notably, heterophilic Wisconsin benefits from higher ER ratios (80-50\% range), likely because ER's unbiased topology provides crucial structural diversity for heterophilic learning, while homophilic Cora shows more robustness to varying ratios. These results show that ER's unbiased topology and cSBM's community structure provide complementary inductive biases essential for universal graph learning.

\begin{table}[ht!]
\centering
\small
\caption{Ablation study on ER/cSBM mixture ratio. }
\label{tab:ablation_ratio}
\begin{tabular}{lccc}
\toprule
ER Ratio & Cora & Wisconsin & Tolokers \\
\midrule
100\% (Only ER) & 80.62 & 80.39 & 77.18 \\
80\% & 80.90 & 81.30 & 77.20 \\
\textbf{50\% (NodePFN)} & \textbf{82.06} & \textbf{81.18} & \textbf{78.61} \\
20\% & 82.01 & 78.75 & 78.10 \\
0\% (Only cSBM) & 81.26 & 78.82 & 77.30 \\
\bottomrule
\end{tabular}
\end{table}

\section{Architectural Ablations}

We provide comprehensive ablation studies on NodePFN's dual-branch architecture to demonstrate the necessity and contribution of each component. Table~\ref{tab:ablation_architecture} shows results for different architectural variants compared to the full NodePFN model.

Removing the MPNN branch causes substantial degradation on both homophilic Cora and heterophilic Wisconsin, demonstrating that the MPNN provides essential structural inductive biases that pure attention cannot capture. NodePFN-Seq underperforms the parallel design. Reducing model capacity to 6 layers (NodePFN-L6) causes failure on Cora, demonstrating that sufficient depth is important for learning diverse patterns from many synthetic priors. These ablations validate that the MPNN provides structural biases, and their parallel combination enables optimal integration, and a deep architecture is required.

\begin{table}[ht!]
\centering
\small
\caption{Ablation study on architectural design choices.}
\label{tab:ablation_architecture}
\begin{tabular}{lccc}
\toprule
Ablation & Cora & Wisconsin & Tolokers \\
\midrule
NodePFN-L6 & 53.10 & 72.94 & 78.00 \\
NodePFN-Seq & 80.64 & 78.82 & 77.88 \\
\midrule
NodePFN w/o MPNN & 75.50 & 70.10 & 78.09 \\
\midrule
\textbf{NodePFN (Full)} & \textbf{82.06} & \textbf{81.18} & \textbf{78.61} \\
\bottomrule
\end{tabular}
\end{table}

\section{Statistics of Synthetic Prior Datasets}
We provide detailed statistics of the synthetic graphs used for pre-training NodePFN. Our training set consists of approximately 250,000 graphs generated over 30 epochs, with each graph sampled from a mixture of ER networks and cSBM with varying homophily ratios.
All synthetic graphs are fixed at 1,024 nodes to balance computational efficiency with sufficient structural complexity for learning meaningful patterns. The number of classes per graph varies from 1 to 20, and edge counts vary based on the underlying generation model and density parameters. On average, each graph contains 12,706.4 edges and 8.79 classes. 

\begin{figure}[h!]
    \centering
    \subfigure[Edge count distribution]{\includegraphics[width=0.49\linewidth]{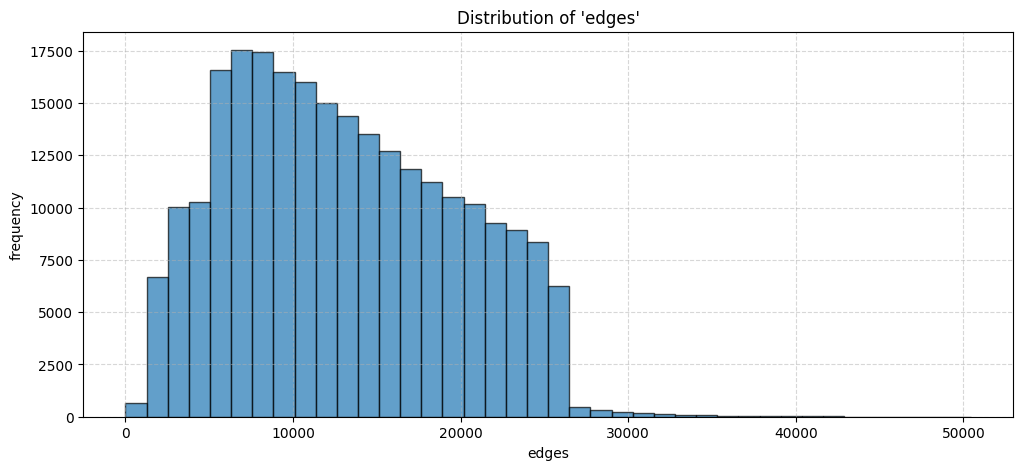}\label{fig:synthetic_distributions-a}}
    \subfigure[Class count distribution]{\includegraphics[width=0.49\linewidth]{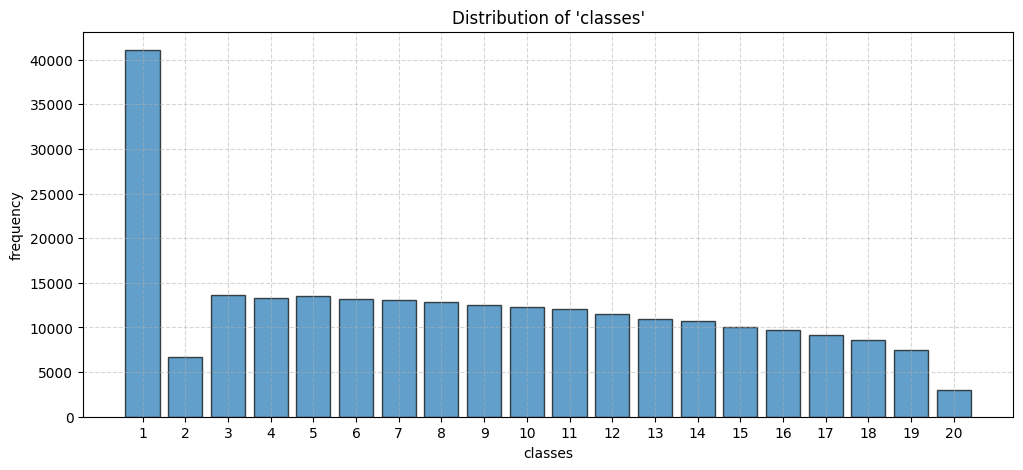}\label{fig:synthetic_distributions-b}}
    \caption{Distribution of synthetic prior datasets used for pre-training.}
    \label{fig:synthetic_distributions}
\end{figure}

\Cref{fig:synthetic_distributions} shows the distributions of edges and classes across all synthetic graphs. The edge distribution (\Cref{fig:synthetic_distributions-a}) exhibits a normal distribution centered around 10,000-15,000 edges, with a long tail extending to sparser graphs. 
This diversity reflects the combination of sparse ER networks, which generate fewer edges on average, and dense cSBM communities, which create higher edge densities within communities.
The class distribution (\Cref{fig:synthetic_distributions-b}) shows coverage across the full range of 1-20 classes, with slightly higher frequency for graphs with fewer classes. 

\section{Computational Complexity}

We provide formal complexity analysis for NodePFN.
For a graph with $N$ nodes, $|E|$ edges, $d$-dimensional features, and $L$ MPNN layers, the MPNN branch requires $\mathcal{O}(L E d)$ operations for message passing and aggregation, identical to standard GCN complexity. The Transformer branch computes attention over all nodes, requiring $\mathcal{O}(N^2  d)$ operations for attention computation and $\mathcal{O}(N d^2)$ for feed-forward layers. The total per-graph complexity is therefore $\mathcal{O}(L E d + N^2 d)$. 

\Cref{tab:runtime} shows comprehensive runtime comparison between GCN and NodePFN across all 23 benchmark datasets. While NodePFN achieves superior average accuracy and ranking, it also demonstrates remarkable computational efficiency in terms of total deployment cost.
GCN requires cumulative training time of 188 seconds plus 12.35 seconds for inference across all datasets.
Importantly, this represents a single training run per dataset with fixed hyperparameters --- in practice, achieving competitive performance typically requires multiple hyperparameter tuning attempts, potentially multiplying this cost by 5 to 10 times or more.
In contrast, NodePFN requires only 47.78 seconds total representing a 4 times speedup in total time-to-deployment.
The one-time pre-training cost (6 GPU hours) is amortized across unlimited datasets, eliminating repetitive per-dataset optimization and making it increasingly efficient as more graphs are processed.

\begin{table}[h!]
\centering
    \small
    \caption{Runtime efficiency comparison}
    \label{tab:runtime}
    \begin{tabular}{l cc c}
    \toprule
    23 Benchmark Datasets &  GCN & NodePFN \\
    \cmidrule(lr){1-3}
     Avg. Accuracy  & 66.63 &  71.27 \\ 
     Avg. Ranking   & 5.86 & 1.70 \\ 
     \cmidrule(lr){1-3}
     Total Train Runtime ($s$)  & 188 & \multirow{2}{*}{47.78} \\ 
     Total Predict Runtime ($s$)  & 12.35 &  \\ 
     \bottomrule
    \end{tabular}
\end{table}

\end{document}